\documentclass{article}
\PassOptionsToPackage{numbers, compress}{natbib}
\usepackage[preprint]{neurips_2026}
\usepackage[utf8]{inputenc} % allow utf-8 input
\usepackage[T1]{fontenc}    % use 8-bit T1 fonts
\usepackage{xcolor}         % colors
\definecolor{cvprblue}{rgb}{0.21,0.49,0.74}
\usepackage[breaklinks,colorlinks,citecolor=cvprblue]{hyperref}  
\usepackage{url}            % simple URL typesetting
\usepackage{booktabs}       % professional-quality tables
\usepackage{amsfonts}       % blackboard math symbols
\usepackage{nicefrac}       % compact symbols for 1/2, etc.
\usepackage{microtype}      % microtypography
\usepackage{graphicx}  
\usepackage{subcaption} 
\usepackage{amsmath}
\usepackage{bm}
\usepackage{tabularx}
\usepackage{multirow}
\usepackage{makecell}
\usepackage{xspace}
\usepackage{longtable}
\usepackage{amsthm}
\usepackage{color, colortbl}
\usepackage{wrapfig}
\usepackage{arydshln}
\usepackage{enumitem}
\usepackage{bbding}
\usepackage[breakable]{tcolorbox}
\definecolor{COLOR_MEAN}{HTML}{f0f0f0}

\definecolor{titlepink}{RGB}{255,105,110}  
\definecolor{rlred}{HTML}{D65551}   
\definecolor{sftyellow}{HTML}{E0AC18} 
\newcommand{\think}[1]{\textcolor{blue}{\texttt{\textbf{<think>}}} #1 \textcolor{blue}{\texttt{\textbf{</think>}}}}
\newcommand{\search}[1]{\textcolor{rlred}{\texttt{\textbf{<tool\_call>}}} #1 \textcolor{rlred}{\texttt{\textbf{</tool\_call>}}}}
\newcommand{\answer}[1]{\textcolor{purple!30}{\texttt{\textbf{<answer>}}} #1 \textcolor{purple!30}{\texttt{\textbf{</answer>}}}} 
\newcommand{\toolresponse}[1]{\textcolor{sftyellow}{\texttt{\textbf{<tool\_response>}}} #1 \textcolor{sftyellow}{\texttt{\textbf{</tool\_response>}}} }

\definecolor{ConBudget}{HTML}{C2185B} % 金钱/预算 - 紫红色
\definecolor{ConAccom}{HTML}{1565C0}  % 住宿/房屋 - 深蓝色
\definecolor{ConTrans}{HTML}{D84315}  % 交通 - 紫色 (偏橙红)
\definecolor{ConRoute}{HTML}{2E7D32}  % 地点/路线 - 深绿色
\definecolor{ConTime}{HTML}{000000}   % 时间/天数 - 黑色/加粗
\definecolor{ConMeal}{HTML}{00838F}   % 餐饮/食物 - 蓝绿色
\newcommand{\cbudget}[1]{\textcolor{ConBudget}{\textbf{#1}}}
\newcommand{\caccom}[1]{\textcolor{ConAccom}{\textbf{#1}}}
\newcommand{\ctrans}[1]{\textcolor{ConTrans}{\textbf{#1}}}
\newcommand{\croute}[1]{\textcolor{ConRoute}{\textbf{#1}}}
\newcommand{\ctime}[1]{\textbf{#1}} 
\newcommand{\cmeal}[1]{\textcolor{ConMeal}{\textbf{#1}}}
\newcommand{\ghlogo}{\raisebox{-0.2\height}{\includegraphics[height=1.6em]{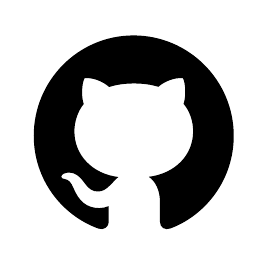}}}
\tcbuselibrary{skins, breakable}

\newtcolorbox{takeawaybox}{
    enhanced,
    breakable,            % 支持跨页
    colback=gray!8,       % 背景色：极浅的灰色 (可以换成 blue!4 变成极浅蓝)
    colframe=gray!8,      % 边框色：与背景一致，实现无边框视觉效果
    arc=3pt,              % 圆角大小
    boxrule=0pt,          % 边框粗细为0
    left=8pt, right=8pt,  % 左右内边距
    top=6pt, bottom=6pt,  % 上下内边距
    before skip=12pt,     % 框上方的外边距
    after skip=12pt       % 框下方的外边距
}

\title{Demystifying Reinforcement Learning for Long-Horizon Tool-Using
Agents:\\ A Comprehensive Recipe}

\author{%
  {
  Xixi Wu$^{1}$\thanks{Contact: \texttt{xxwu@se.cuhk.edu.hk}. Corresponding authors: \texttt{\{qiyiyan@idea.edu.cn, hcheng@se.cuhk.edu.hk\}}} \hspace{0.4cm}
  Qianguo Sun$^{2}$\hspace{0.4cm}
  Ruiyang Zhang$^{2,3}$ \hspace{0.4cm}
  Chao Song$^2$
  }
 \vspace{0.1cm}
 \\
 {
  \textbf{Junlong Wu}$^2$\hspace{0.4cm}
  \textbf{Yiyan Qi}$^2$ \Envelope\hspace{0.4cm}
  \textbf{Hong Cheng}$^1$ \Envelope
  }
  \vspace{0.35cm}
  \\
  { \normalfont $^1$The Chinese University of Hong Kong\hspace{0.16cm} $^2$IDEA Research}\\
  $^3$University of Macau\\
 {{\color{black}\ghlogo~}\href{https://github.com/WxxShirley/Agent-STAR}{\texttt{https://github.com/WxxShirley/Agent-STAR}}}
}

\begin{document}

\maketitle

\begin{abstract}
 Reinforcement Learning (RL) is essential for evolving Large Language Models (LLMs) into autonomous agents capable of long-horizon planning, yet a practical recipe for scaling RL in complex, multi-turn environments remains elusive. This paper presents a systematic empirical study using TravelPlanner, a challenging testbed requiring tool orchestration to satisfy multifaceted constraints. We decompose the agentic RL design space along 5 axes: reward shaping, model scaling, data composition, algorithm selection, and environmental stability. Our controlled experiments yield 7 key takeaways, e.g., (1) reward and algorithm choices are scale-dependent as smaller models benefit from staged rewards and enhanced exploration, whereas larger models converge efficiently with simpler dense rewards, (2) $\sim$1K training samples with a balanced difficulty mixture mark a sweet spot for both in-domain and out-of-domain performance, and (3) environmental stability is critical to prevent policy degradation. Based on our distilled recipe, our RL-trained models achieve state-of-the-art performance on TravelPlanner, significantly outperforming leading LLMs. 
\end{abstract}

\begin{figure}[!h]
    \centering
    \includegraphics[width=0.55\linewidth]{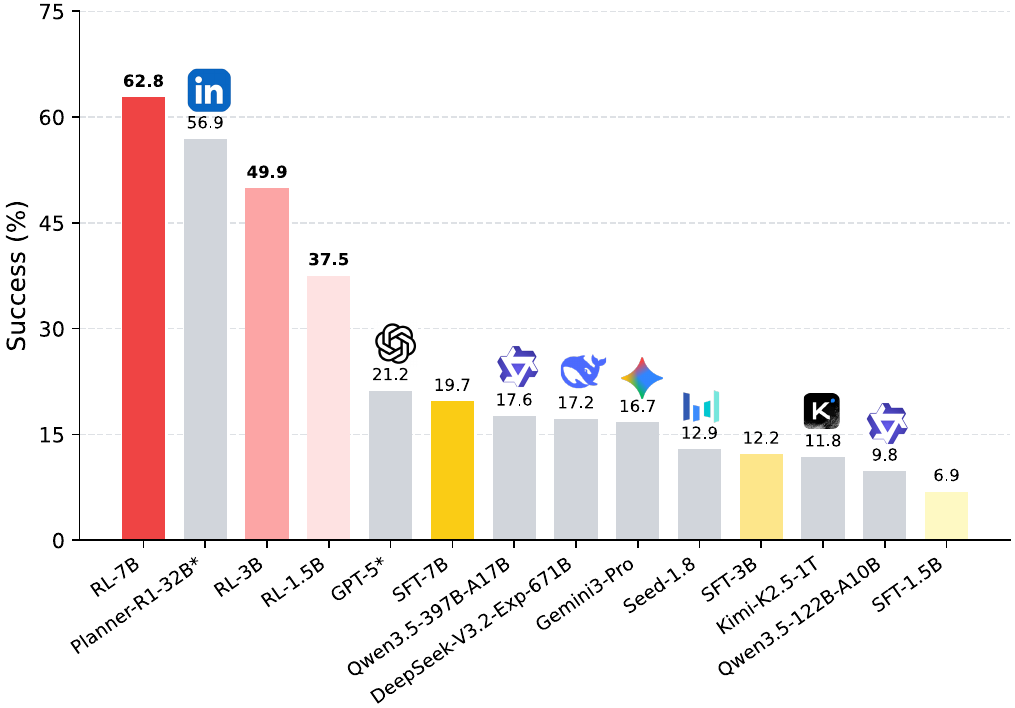}
    \vspace*{-5pt}
    \caption{\textbf{Performance on the TravelPlanner test set.} Our \textbf{\textcolor{rlred}{RL-trained models}} significantly outperform both their \textbf{\textcolor{sftyellow}{SFT counterparts}} and \textbf{\textcolor{gray}{leading LLMs}}. Results marked with $^*$ are sourced from~\citet{zhu2025plannerr1}.}
    \label{fig:sota_test_performance}
    \vspace*{-10pt}
\end{figure}

\section{Introduction}
Large Language Models (LLMs) have evolved from static text generators into general-purpose autonomous agents capable of reasoning, acting, and interacting with dynamic environments \cite{yao2023react, schick2023toolformer}. This paradigm shift has enabled diverse applications, ranging from information-seeking agents navigating open-ended web environments~\cite{jin2025search, li2025websailor} to GUI agents manipulating complex user interfaces~\cite{OSWorld} and software engineering agents modifying and debugging real-world codebases~\cite{jimenez2024swebench, yang2024swebenchmultimodal, yang2024sweagent}. Across these scenarios, agents must engage in long-horizon planning: decomposing high-level goals into manageable sub-tasks, orchestrating tool usage, and satisfying multifaceted constraints to ensure the successful completion of tasks~\cite{wu2024graph, xie2024travelplanner}.

Training agents capable of long-horizon tool use remains an open challenge, establishing Reinforcement Learning (RL) as a primary paradigm for optimizing these capabilities through exploration and feedback~\cite{kimi-researcher2025,li2025websailorv2}. However, existing insights into agentic RL stem predominantly from short-horizon tasks involving single-step reasoning~\cite{yeotong2025longcot} or few-turn interactions~\cite{jin2025empirical,yu2025demystify}. In contrast, real-world agentic workflows require long-horizon planning, characterized by dozens of tool invocations and extensive trajectories. While recent efforts have introduced targeted algorithms to tackle this complexity, such as modifying exploration strategies~\cite{ji2025treegrpo,dong2025arpo} or synthesizing adaptive environments~\cite{lu2025envtune, wang2026rlanything}, these works typically explore a limited subset of the RL design space. Crucially, they lack a holistic view of how factors ranging from reward shaping and data composition to model scaling and environmental stability jointly shape performance. Therefore, the community still lacks a comprehensive and practical recipe for scaling RL in complex, long-horizon agentic scenarios.

To fully explore this design space and bridge the aforementioned gap, we require an environment that is both complex yet computationally tractable. We adopt \textbf{TravelPlanner}~\cite{xie2024travelplanner} as our primary testbed, which perfectly exemplifies the challenges of long-horizon agents. It requires \textbf{orchestrating diverse tools} (e.g., transport and accommodation search) to satisfy \textbf{multifaceted constraints} (e.g., budget, personal preferences, and hallucination avoidance), presenting a challenge where even top-tier models such as Kimi-K2.5~\cite{kimiteam2026kimik25} achieve success rates below 15\%. Unlike tasks relying on costly and high-latency external APIs, TravelPlanner operates within a local sandbox, providing the zero-cost, high-throughput simulation essential for scaling RL exploration. Leveraging this efficient testbed, we implement \textbf{STAR} (\underline{S}ynthesis, \underline{T}raining, \underline{A}nd \underline{R}einforcement), a unified post-training pipeline designed to systematically instill and refine long-horizon planning capabilities. Furthermore, moving beyond intra-task evaluation, we assess our trained policies on both in-domain planning tasks and out-of-domain (OOD) knowledge-intensive QA benchmarks to evaluate their broader generalization.

Utilizing the STAR framework, we conduct a large-scale empirical study to decompose the long-horizon RL design space along 5 critical axes: \textbf{reward shaping} (dense \textit{vs}. sparse, with or without curriculum-style staging), \textbf{model scaling} (1.5B, 3B, and 7B variants), \textbf{data composition} (sample quantity and difficulty), \textbf{algorithm selection} (standard GRPO \textit{vs}. exploration-heavy variants), and \textbf{environmental stability} (injecting random tool failures). By rigorously isolating each factor, we distill the following key takeaways: (1) \textbf{Reward and algorithm choices are scale-dependent:} smaller models benefit most from staged curriculum rewards and exploration-heavy algorithms, whereas larger models favor simpler dense rewards and standard GRPO for both accuracy and efficiency; (2) \textbf{Data exhibits a sweet spot:} approximately 1K training samples with a balanced difficulty mixture provide the optimal trade-off between in-domain performance and OOD generalization; and (3) \textbf{Environmental stability is critical:} environmental noise can noticeably degrade the performance of long-horizon agents. Finally, following the optimal strategies identified across these factors, our STAR-trained 1.5B-7B models achieve state-of-the-art (SOTA) performance on the TravelPlanner test set, significantly outperforming the strongest commercial LLMs as shown in Figure~\ref{fig:sota_test_performance}.

In summary, our contributions are as follows:
\begin{itemize}
    \item \textbf{A Holistic Post-training Pipeline:} We leverage TravelPlanner as a scalable testbed for long-horizon agents and develop STAR, a unified pipeline encompassing data synthesis, supervised fine-tuning (SFT), and RL, validated across both in-domain and OOD tasks.
    \item \textbf{A Large-scale Empirical Study:} We systematically dissect the RL design space, providing empirical insights into how reward shaping, model scaling, data composition, algorithm selection, and environmental stability jointly determine policy optimization.
    \item \textbf{Actionable Recipe \& SOTA Performance:} We derive a practical, scale-aware recipe for training long-horizon agents. Applying this recipe, our open-weight models achieve SOTA performance on TravelPlanner, surpassing leading proprietary LLMs and providing a foundation for future agentic RL research. 
\end{itemize}

\section{Preliminaries}
\begin{figure*}[!h]
    \centering
    \includegraphics[width=0.95\linewidth]{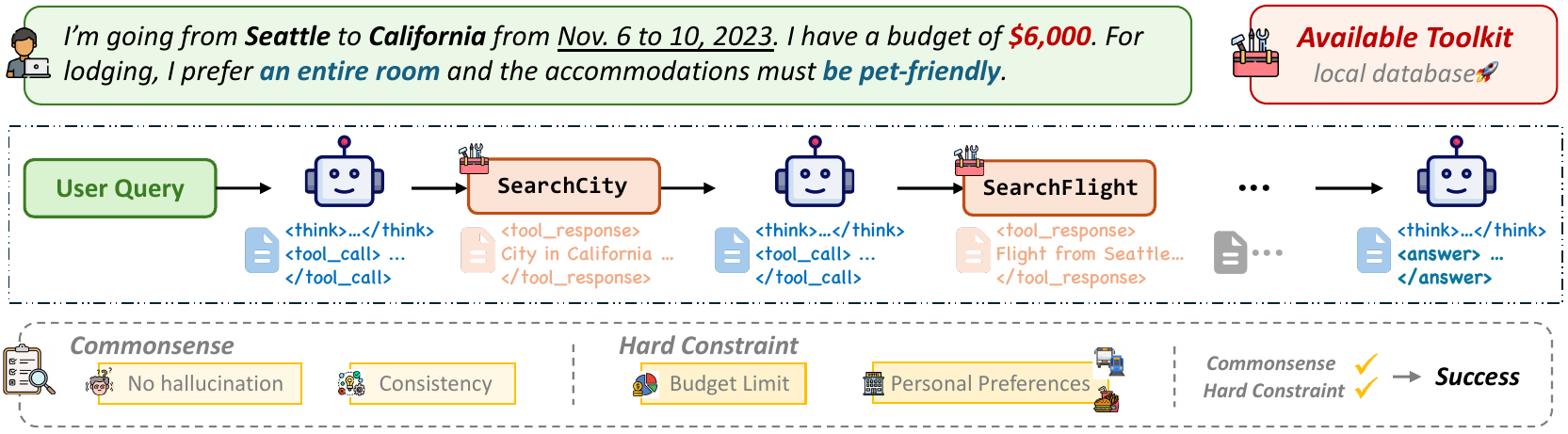}
    \caption{\textbf{Overview of the TravelPlanner testbed.} \textbf{Top:} A user query alongside the available local toolkit. \textbf{Middle:} The agent's iterative reasoning and tool-use trajectory following the ReAct paradigm. \textbf{Bottom:} The two-dimensional evaluation protocol, requiring adherence to both commonsense rules and hard constraints to achieve a \textit{\textbf{Success}}.}
    \label{fig:travelplanner_workflow}
\end{figure*}

We use \textbf{TravelPlanner}~\cite{xie2024travelplanner} as our primary testbed. This platform simulates a realistic travel agency scenario where agents execute long-horizon planning under multifaceted constraints. These constraints encompass both \textit{explicit user requirements} (e.g., budget limits and personal preferences) and \textit{implicit commonsense rules} (e.g., factual grounding and logical consistency). The testbed provides 6 information-gathering tools (e.g., \texttt{SearchFlight}) that query a large-scale local database, with detailed statistics provided in \textbf{Appendix~\ref{appendix:travelplanner_detail}} and Table~\ref{tab:travelplanner_sandbox}. This configuration replicates the complexity of real-world APIs while ensuring the zero-cost, low-latency interactions essential for scalable RL.

\noindent \textbf{ReAct Inference:} As illustrated in Figure~\ref{fig:travelplanner_workflow}, we employ the ReAct paradigm~\cite{yao2023react} to facilitate multi-turn agentic workflows. Given a natural language query \(q\) specifying the travel intent and constraints, the agent engages in iterative cycles of reasoning, acting, and observing. At each time step \(t\), the LLM generates a reasoning trace \(\tau_t\) conditioned on the context, emits a parsable tool action \(a_t\), and receives an observation \(o_t\) from the testbed. The process terminates when the agent produces a final natural language itinerary \(a_T\), yielding a complete trajectory defined as:
\[
\mathcal{H}_T = (q, \tau_1, a_1, o_1, \ldots, \tau_{T}, a_T).
\]

\noindent \textbf{Evaluation Protocol:} Given the unstructured nature of the final natural language plan \(a_T\), we employ a dedicated formatting model to parse the output into a structured JSON itinerary prior to automated evaluation, as detailed in Appendix~\ref{appendix:plan_formatting}. We evaluate performance along two dimensions, with specific rules outlined in Table~\ref{tab:evaluation_rules}: Commonsense (denoted as $\text{cs}$, e.g., logical consistency, absence of hallucinations) and Hard Constraints (denoted as $\text{hard}$, e.g., adherence to budget and dietary restrictions). For each dimension $k \in \{\text{cs}, \text{hard}\}$, we compute a \textbf{micro score} $s_k^{\text{micro}} \in [0,1]$, representing the ratio of satisfied checks, and a binary \textbf{macro score} $s_k^{\text{macro}} = \mathbb{I}(s_k^{\text{micro}} = 1)$, indicating full compliance. A trajectory is deemed \textbf{\textit{Success}} if and only if all constraints are met as follows:
\begin{equation*}
    s^{\text{success}} = \mathbb{I}\left(s_{\text{cs}}^{\text{macro}} = 1 \land s_{\text{hard}}^{\text{macro}} = 1\right).
\end{equation*}

\begin{figure*}[!t]
    \centering
    \includegraphics[width=0.96\linewidth]{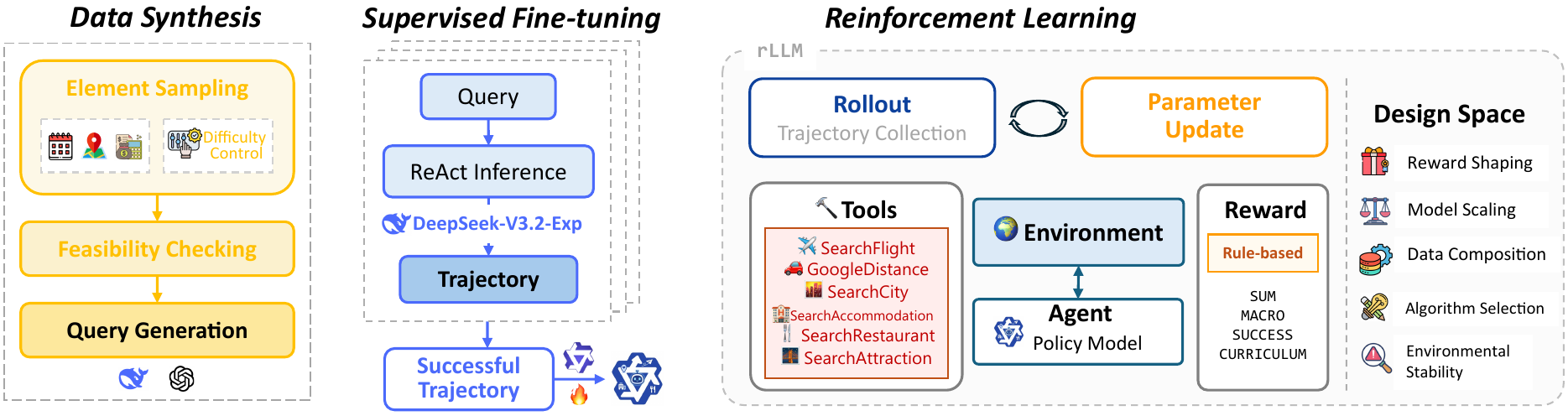}
    \caption{\textbf{Overview of the STAR pipeline and the experimental design space.} The pipeline consists of three stages: (1) \textbf{Data Synthesis} for generating feasible, difficulty-controlled queries, (2) \textbf{SFT} using high-quality trajectories from a teacher model, and (3) \textbf{RL} to optimize long-horizon planning via environmental feedback. The right panel outlines the design space explored in our subsequent experiments.}
    \label{fig:star_workflow}
\end{figure*}

\section{STAR Pipeline}
We introduce the \textbf{STAR} pipeline, a unified post-training framework designed for long-horizon agents on TravelPlanner. As illustrated in Figure~\ref{fig:star_workflow}, the pipeline comprises three sequential stages: data synthesis to construct queries with controllable difficulty, SFT to obtain task-aware initial policies, and RL to further strengthen long-horizon planning behaviors. 

\noindent \textbf{Data Synthesis:} Addressing the scarcity of training data, we develop a synthesis procedure to generate additional TravelPlanner-style queries. We first sample atomic travel elements (e.g., origin, destination, and dates) and validate their feasibility within the sandbox to ensure the existence of ground-truth solutions. Using these validated constraints and dynamically estimated budgets, we employ open-sourced models~\cite{deepseekai2024deepseekv32exp,openai2025gptoss120bgptoss20bmodel} to generate natural language queries via back-translation. To obtain queries with controllable difficulty, we follow the original TravelPlanner design, categorizing them into specific difficulty levels (i.e., easy, medium, and hard) by varying the number and types of constraints. Detailed definitions and concrete examples of difficulty levels are provided in \textbf{Appendix Table~\ref{tab:query_difficulty}}.

\noindent \textbf{SFT:} To mitigate the cold-start issue and equip the policy with basic task understanding, we apply SFT prior to RL. We follow a rejection-sampling style procedure: first selecting a strong teacher model to perform ReAct inference on the synthesized queries, retaining only trajectories that achieve \textit{\textbf{Success}} under the evaluation protocol. The resulting high-quality trajectories serve as gold supervision for SFT, yielding task-specialized initial checkpoints for all model sizes.

\noindent \textbf{RL:} The core of our framework is the RL stage, where the agent optimizes long-horizon planning via environmental feedback. We utilize \texttt{rLLM}~\cite{rllm2025}, a popular framework for post-training language agents. Aligned with the evaluation protocol, we implement a spectrum of reward signals ranging from dense to sparse:
\begin{itemize}
\item \textbf{\textsc{Sum:}} A dense reward aggregating all sub-metrics, defined as $r^{\text{sum}} = s_{\text{cs}}^{\text{micro}} + s_{\text{cs}}^{\text{macro}} + s_{\text{hard}}^{\text{micro}} + s_{\text{hard}}^{\text{macro}} + s^{\text{success}}$.
\item \textbf{\textsc{Macro:}} A semi-sparse reward focusing on macro-level constraint satisfaction, defined as $r^{\text{macro}} = s_{\text{cs}}^{\text{macro}} + s_{\text{hard}}^{\text{macro}} + s^{\text{success}}$.
\item \textbf{\textsc{Success:}} A purely sparse binary reward, defined as $r^{\text{success}} = s^{\text{success}}$.
\item \textbf{\textsc{Curriculum:}} Following~\citet{zhu2025plannerr1}, we implement a staged curriculum where the reward function transitions from $r^{\text{sum}}$ to $r^{\text{success}}$ during training to guide exploration.
\end{itemize}

We employ GRPO~\cite{shao2024deepseekmath} as the primary optimization algorithm. For a query $q$, we sample a group of $G$ trajectories $\{\mathcal{H}^{(1)}, \ldots, \mathcal{H}^{(G)}\}$ from the old policy $\pi_{\theta_{\text{old}}}$. The objective maximizes the surrogate advantage as follows:
\begin{equation*}
\small
\label{eq:grpo}
\begin{aligned}
    \mathcal{J}_{\text{GRPO}}(\theta) = &  \mathbb{E}_{q \sim \mathcal{D}, \{ \mathcal{H}^{(i)}\}_{i=1}^G \sim \pi_{\theta_{\text{old}}}} \\ & \Bigg[ \frac{1}{\sum_{i=1}^{G} |\mathcal{H}^{(i)}|} \sum_{i=1}^G  \sum_{j=1}^{|\mathcal{H}^{(i)}|} \min \Big(  \rho_{j}^{(i)} \hat{A}^{(i)}, \text{clip}(\rho_{j}^{(i)}, 1-\epsilon_{\text{low}}, 1+\epsilon_{\text{high}}) \hat{A}^{(i)} \Big) \Bigg],
\end{aligned}
\end{equation*}
where $\rho_{j}^{(i)}$ is the importance sampling ratio, \(\epsilon_{\text{low}}\) and \(\epsilon_{\text{high}}\) denote the asymmetric clipping bounds, and $\hat{A}^{(i)}$ is the advantage computed by normalizing rewards within the sampled group. Finally, to systematically explore the RL design space, we extend \texttt{rLLM} into a modular setup that flexibly varies data, rewards, algorithms, and environmental dynamics, facilitating subsequent empirical study.

\begin{table}[!t]
\centering
\caption{\textbf{In-domain performance (\%) across reward designs on the TravelPlanner test set.} Best results for each model scale are in \textbf{bold}. Smaller models benefit most from the progressive \textsc{Curriculum} reward, while the stronger 7B model favors the dense \textsc{Sum} reward.}
\label{tab:reward_indomain}
\vspace*{2pt}
\resizebox{0.7\linewidth}{!}{
\begin{tabular}{ccccccc}
\toprule
\rowcolor{COLOR_MEAN} &  & \multicolumn{2}{c}{\textbf{Commonsense}} & \multicolumn{2}{c}{\textbf{Hard Constraint}} & \\
\rowcolor{COLOR_MEAN}  \multirow{-2}{*}{\textbf{Scale}} & \multirow{-2}{*}{\textbf{Method}} & \textbf{Micro} & \textbf{Macro} & \textbf{Micro} & \textbf{Macro} & \multirow{-2}{*}{\textbf{Success}} \\ 

\midrule
\multirow{6}{*}{{\textbf{1.5B}}} 
& Base & 30.1 & 0.0 & 0.0 & 0.0 & 0.0 \\
& SFT & 65.9 & 15.4 & 17.2 & 12.1 & 6.9 \\
\cmidrule{2-7}
& \textsc{Sum} & \textbf{95.1} & \textbf{71.6} & \textbf{51.4} & 33.4 & 33.1 \\
& \textsc{Macro} & 93.4 & 68.4 & 47.9 & 31.6 & 30.1 \\
& \textsc{Success} & 93.9 & 68.2 & 51.0 & 34.7 & 33.8 \\
& \small{\textsc{Curriculum}} & 93.9 & 70.2 & 51.0 & \textbf{35.4} & \cellcolor{pink!30}\textbf{34.9} \\
\midrule
\multirow{6}{*}{{\textbf{3B}}} 
& Base & 46.5 & 0.0 & 0.0 & 0.0 & 0.0  \\ 
& SFT & 70.4 & 24.6 & 28.6 & 20.0 & 12.2 \\
\cmidrule{2-7}
& \textsc{Sum} & \textbf{97.6} & 82.5 & 64.8 & 47.6 & 47.0 \\
& \textsc{Macro} & 95.1 & 79.6 & \textbf{68.8} & \textbf{52.3} & 48.2 \\
& \textsc{Success} & 90.6 & 76.1 & 62.6 & 47.7 & 46.6 \\
& \small{\textsc{Curriculum}} & 95.3 & \textbf{83.0} & 67.6 & 51.0 & \cellcolor{pink!30}\textbf{49.9} \\
\midrule
\multirow{6}{*}{{\textbf{7B}}} 
& Base  & 55.8 & 0.1 & 0.7 & 0.7 & 0.1 \\
& SFT & 77.1 & 33.1 & 40.6 & 31.3 & 19.7 \\
\cmidrule{2-7}
& \textsc{Sum} & 96.9 & 87.4 & \textbf{78.7} & \textbf{66.5} & \cellcolor{pink!30}\textbf{62.8} \\
& \textsc{Macro} & \textbf{97.0} & \textbf{89.8} & 77.3 & 60.0 & 58.9 \\
& \textsc{Success} & 91.4 & 78.3 & 73.4 & 64.0 & 60.9 \\
& \small{\textsc{Curriculum}} & 96.1 & 85.1 & 76.6 & 60.4 & 57.0 \\
\bottomrule
\end{tabular}
}
\vspace*{-10pt}
\end{table}

\section{Experiments}
\subsection{Setup} 
\noindent \textbf{Pipeline Instantiation:} We instantiate the three-stage STAR pipeline with strict quality controls to ensure a rigorous testbed.
\begin{itemize}
\item \textbf{Data Synthesis:} We synthesize over 10K queries with a balanced difficulty ratio using strong open-weight models, including GPT-OSS-120B~\cite{openai2025gptoss120bgptoss20bmodel} and DeepSeek-V3.2-Exp~\cite{deepseekai2024deepseekv32exp}. To verify data reliability, we evaluate 200 sampled synthetic queries and confirm that their success rate closely aligns with that of the official TravelPlanner validation set.

\item \textbf{SFT:} We prompt DeepSeek-V3.2-Exp-Thinking on 5K synthetic queries to perform ReAct inference. Filtering strictly for task \textit{\textbf{Success}} and format adherence yields \textbf{1,198 high-quality trajectories} that average 10.3K tokens and 9.2 tool calls, as detailed in Appendix Table~\ref{tab:sft_data}. We use these to fine-tune the \textbf{Qwen2.5-Instruct} series~\cite{qwen2025qwen25technicalreport} as our SFT base. We intentionally restrict the scale of this SFT phase to establish protocol adherence without inducing policy collapse, thereby preserving exploration space for the subsequent RL stage.

\item \textbf{RL:} We employ GRPO with practical modifications following~\citet{yu2025dapo} to stabilize training: (1) KL-Free \& Clip-high: We remove the KL penalty and increase the clipping bound $\epsilon_{\text{high}}$ to encourage broader exploration. (2) Strict protocol enforcement: Trajectories with format errors receive a reward of 0. (3) Overlength handling: To prevent instability, overlength rollouts are excluded from loss computation but retained for advantage normalization to maintain statistical robustness, following~\citet{zhao2026repurposing}.
\end{itemize}

\noindent \textbf{Default Configurations:}
Unless otherwise specified, our default RL training uses \textbf{1K synthetic queries}, ensuring no overlap with the SFT data, with a 4:3:3, easy:medium:hard, difficulty ratio. Models are trained for 5 epochs with a group size \(G=8\). The maximum context length is set to 30K tokens during training and extended to 32K for inference. Model selection relies on the TravelPlanner validation set. We conduct controlled experiments by strictly varying one factor at a time while keeping others fixed.

\noindent \textbf{Evaluation:}
We evaluate in-domain performance on the 1,000-instance TravelPlanner test set. For OOD generalization, we report results on 7 distinct knowledge-intensive QA benchmarks, comparing against strong domain-specific baselines. Following~\citet{jin2025search}, the only available tool for these OOD tasks is a local Wikipedia search engine. Due to space limits, further implementation details are deferred to \textbf{Appendix~\ref{appendix:implementation}}.

\subsection{Reward Shaping}
\textbf{Motivation:} A critical open question in RL for long-horizon agents is how the density of reward signals impacts reasoning capabilities. To answer this question, we evaluate a spectrum of reward designs ranging from dense \textbf{\textsc{Sum}} and semi-sparse \textbf{\textsc{Macro}}, to purely sparse \textbf{\textsc{Success}}. Furthermore, we evaluate a \textbf{\textsc{Curriculum}} reward~\cite{zhu2025plannerr1} that progressively transitions from dense to sparse. This acts as a staged intervention based on human priors, providing fine-grained guidance during the early training phases. To strictly isolate the effect of reward shaping, all RL configurations share identical training data, base models, and hyperparameters, as detailed in Appendix~\ref{appendix:reward}.

\begin{table*}[!t]
\centering
\caption{\textbf{Out-of-domain performance (in \%) on the knowledge-intensive QA benchmarks}. Results marked with $^\star$ are sourced from~\citet{ji2025treegrpo}. The best results of each dataset across methods are highlighted in \textbf{bold}. Our RL-trained models achieve comparable or superior performance to the domain-specific Search-R1 baseline~\cite{jin2025search} across model scales, showing that complex tool-calling capabilities effectively generalize to other domains.}
\label{tab:reward_ood}
\vspace*{-5pt}
\resizebox{\linewidth}{!}{
\begin{tabular}{cccccccccc}
\toprule
\rowcolor{COLOR_MEAN} \textbf{Scale} & \textbf{Method} & \textbf{NQ} & \textbf{TriviaQA} & \textbf{PopQA} & \textbf{HotpotQA} & \textbf{2Wiki} & \textbf{Musique} & \textbf{Bamboogle} & \textbf{Avg.} \\
\midrule
\multirow{6}{*}{\textbf{1.5B}} 
& Base$^\star$  & 7.1 & 22.4 & 9.9 & 5.9 & 4.3 & 2.6 & 8.0 & 8.6 \\
& SFT & 13.5 & 21.4 & 13.5 & 11.6 & 7.0 & 1.2 & 12.0 & 11.5 \\
& Search-R1$^\star$ & \textbf{39.4} & \textbf{51.0} & \textbf{39.7} & 14.6 & \textbf{24.4} & 2.2 & 4.0 & 25.0 \\ 
\cmidrule{2-10} 
& \textsc{Sum} & 32.1 & 44.5 & 30.9 & 26.0 & 14.8 & 5.3 & 16.8 & 24.3 \\
& \textsc{Macro} & 29.6 & 45.1 & 28.8 & \textbf{26.9} & 19.4 & \textbf{6.3} & \textbf{23.2} & \cellcolor{pink!30}\textbf{25.6} \\
& \textsc{Success} & 24.8 & 36.9 & 21.3 & 24.2 & 17.7 & 4.6 & 23.2 & 21.8 \\
& \textsc{Curriculum} & 20.6 & 28.5 & 20.5 & 16.0 & 11.7 & 3.5 & 17.6 & 16.9 \\
\midrule
\multirow{7}{*}{\textbf{3B}} 
& Base & 10.6 & 28.8 & 10.8 & 14.9 & 24.4 & 2.0 & 2.4 & 13.4 \\
& SFT & 35.1 & 52.5 & 31.3 & 32.0 & 24.4 & 9.0 & 30.4 & 30.7 \\
& Search-R1 & 34.1 & 54.5 & \textbf{37.8} & 32.4 & \textbf{31.9} & 10.3 & 26.4 & 32.5 \\
\cmidrule{2-10}  
& \textsc{Sum} & 38.2 & 54.9 & 34.3 & 34.2 & 23.3 & 8.8 & 27.2 & 31.6 \\
& \textsc{Macro} & 37.0 & 54.5 & 34.9 & 37.9 & 29.9 & 11.6 & \textbf{33.6} & 34.2 \\
& \textsc{Success} & 37.5 & 54.1 & 34.0 & 33.7 & 19.4 & 10.3 & 32.0 & 31.6 \\
& \textsc{Curriculum} & \textbf{41.0} & \textbf{56.8} & 36.2 & \textbf{39.5} & 27.7 & \textbf{12.4} & 32.0 & \cellcolor{pink!30}\textbf{35.0} \\
\midrule
\multirow{7}{*}{\textbf{7B}} 
& Base & 13.4 & 40.8 & 14.0 & 18.3 & 25.0 & 3.1 & 12.0 & 18.1 \\
& SFT & 41.2 & 59.7 & 37.6 & 47.0 & 37.8 & 16.3 & \textbf{53.6} & 41.9 \\
& Search-R1 & 39.3 & 61.0 & \textbf{39.7} & 37.0 & \textbf{41.4} & 14.6 & 36.8 & 38.5 \\
\cmidrule{2-10} 
& \textsc{Sum} & 35.5 & 54.5 & 34.8 & 44.8 & 34.6 & 15.4 & 37.6 & 36.7 \\
& \textsc{Macro} & \textbf{42.2} & \textbf{61.2} & 39.6 & \textbf{48.8} & 38.3 & \textbf{17.4} & 52.8 & \cellcolor{pink!30}\textbf{42.9} \\
& \textsc{Success} & 38.7 & 55.9 & 35.2 & 45.8 & 38.4 & 15.5 & 41.6 & 38.7 \\
& \textsc{Curriculum} & 41.1 & 58.7 & 37.7 & 48.4 & 38.5 & 17.4 & 45.6 & 41.1 \\
\bottomrule
\end{tabular}
}
\end{table*}

\noindent \textbf{Table~\ref{tab:reward_indomain}} presents the in-domain performance on TravelPlanner. We compare our RL variants against two baselines: the pre-trained Base models and their SFT counterparts, which serve as the starting checkpoints for RL. For a comprehensive analysis, training dynamics and OOD generalization results are provided in \textbf{Figure~\ref{fig:reward_training_dynamics}} in Appendix~\ref{appendix:reward} and \textbf{Table~\ref{tab:reward_ood}}, respectively. Synthesizing these results yields two takeaways.

% \noindent \textbf{Takeaway 1: Reward design is scale-dependent; sparse-only rewards are suboptimal.} 
\begin{takeawaybox}
\textbf{Takeaway 1: Reward design is scale-dependent; sparse-only rewards are suboptimal.} 
\end{takeawaybox}
In the TravelPlanner domain, smaller models struggle with credit assignment over long horizons and benefit significantly from staged guidance. Consequently, the \textsc{Curriculum} reward achieves the highest success rates and accelerates convergence, as shown in Figure~\ref{fig:reward_training_dynamics}. Conversely, the stronger 7B model possesses the intrinsic capacity to directly leverage fine-grained feedback from the dense \textsc{Sum} reward, rendering heuristic staged transitions unnecessary and even slightly restrictive. Notably, while the sparse \textsc{Success} reward is competitive, it never achieves the best performance across any scale, indicating that outcome-only feedback is insufficient for optimizing long-horizon trajectories.

% \noindent \textbf{Takeaway 2: Overly dense rewards induce an alignment tax on OOD generalization.} 
\begin{takeawaybox}
\textbf{Takeaway 2: Overly dense rewards induce an alignment tax on OOD generalization.} 
\end{takeawaybox}
While the \textsc{Sum} reward maximizes in-domain performance for the 7B model, Table~\ref{tab:reward_ood} reveals a severe alignment tax: its average OOD accuracy falls significantly behind the SFT checkpoint. This indicates that overly dense, task-specific rewards cause the model to overfit to the TravelPlanner format, degrading its general information-seeking abilities. Conversely, the semi-sparse \textsc{Macro} reward achieves an optimal balance, preserving generalization capabilities while remaining highly competitive in-domain.

\subsection{Model Scaling} 

\textbf{Motivation:} Beyond reward design, a fundamental question is whether scaling model capacity inherently resolves the reasoning bottlenecks in long-horizon RL. To investigate this, we compare the 1.5B, 3B, and 7B models under fixed reward configurations. This allows us to evaluate if larger underlying architectures are  better equipped to handle the complexities of multi-turn tool-use and planning.

\begin{wrapfigure}{r}{0.5\textwidth}
\centering
\includegraphics[width=\linewidth]{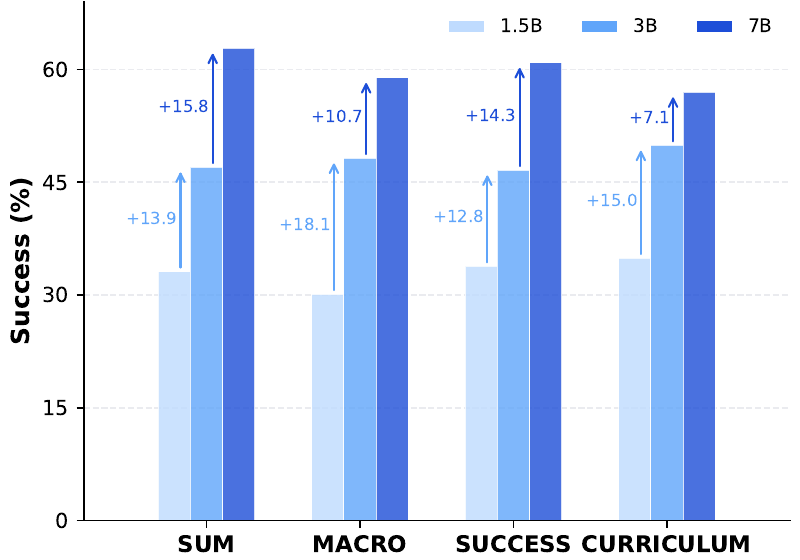}
\caption{\textbf{Performance across model scales on the TravelPlanner test set.} Scaling up model capacity consistently improves success rates across all reward types, though the magnitude of improvement varies.}
\label{fig:modelscale_test}
\vspace*{-2.2cm}
\end{wrapfigure}

\noindent \textbf{Figure~\ref{fig:modelscale_test}} illustrates the in-domain success rates on the TravelPlanner test set across different model scales. Synthesizing these results, along with the training dynamics shown in \textbf{Figure~\ref{fig:modelscale_training_dynamics}} in Appendix~\ref{appendix:model_scale}, reveals a clear scaling behavior.

% \noindent \textbf{Takeaway 3: Scaling model size consistently improves long-horizon performance, but the magnitude of gains varies across reward designs.} 
\begin{takeawaybox}
\textbf{Takeaway 3: Scaling model size consistently improves long-horizon performance, but the magnitude of gains varies across reward designs.} 
\end{takeawaybox}
As shown in Figure~\ref{fig:modelscale_test}, transitioning from the 1.5B to the 7B architecture yields substantial improvements in success rates across all reward signals. For instance, under the dense \textsc{Sum} reward, the success rate nearly doubles from 33.1\% at 1.5B to 62.8\% at 7B. This upward trend is further corroborated by the training dynamics in Figure~\ref{fig:modelscale_training_dynamics}, which demonstrate that larger models not only converge faster but also reach significantly higher performance asymptotes. While scaling is universally beneficial, we observe that the specific rate of improvement is reward-dependent, e.g., moving from 3B to 7B yields a 15.8\% absolute gain under \textsc{Sum}, \textit{vs.} only 7.1\% under \textsc{Curriculum}. Ultimately, these findings indicate that base model capacity remains a primary bottleneck for complex agentic tasks, and that RL effectively unlocks these inherent reasoning capabilities, particularly when guided by suitable reward designs.

\subsection{Data Composition}

\begin{figure}[htbp]
\begin{minipage}[c]{0.48\textwidth}
\centering
 \includegraphics[width=\textwidth]{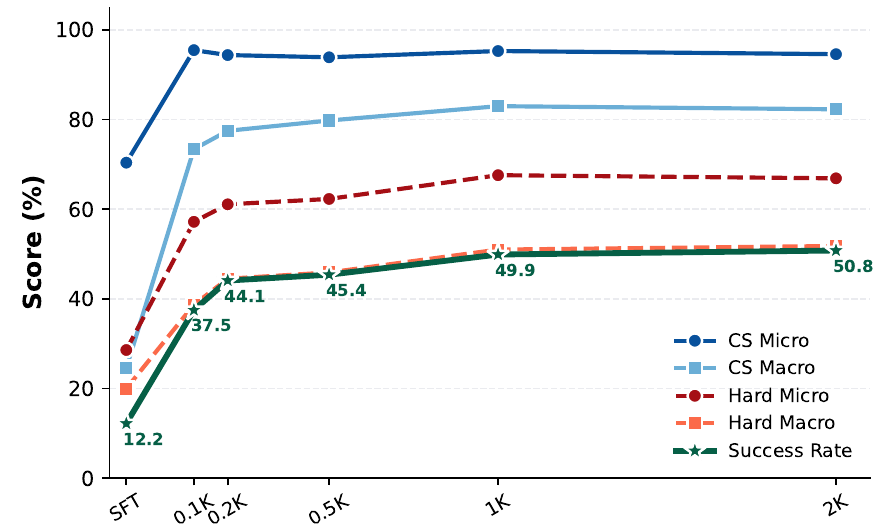}
\caption{\textbf{Performance across different training data quantities on the TravelPlanner test set.} A sweet spot emerges at 1K prompts, balancing in-domain success with strong OOD generalization, which is detailed in Appendix Table~\ref{tab:data_ood}.}
\label{fig:data_quantity_test}
\end{minipage}\hfill
\begin{minipage}[c]{0.5\textwidth}
\centering
\captionof{table}{\textbf{In-domain performance (\%) across data difficulty compositions on the TravelPlanner test set.} Models are trained on 1K prompts from the same SFT-3B checkpoint using the \textsc{Curriculum} reward. The notation XX-1K denotes the data composition, e.g., Easy-1K contains exclusively easy queries. \textbf{Mixed-1K} blends easy:medium:hard at a 4:3:3 ratio.}
\label{tab:data_difficulty_indomain}
\resizebox{\linewidth}{!}{
\begin{tabular}{cccccc}
\toprule
\rowcolor{COLOR_MEAN}  & \multicolumn{2}{c}{\textbf{Commonsense}} & \multicolumn{2}{c}{\textbf{Hard Constraint}} & \\
\rowcolor{COLOR_MEAN}  \multirow{-2}{*}{\textbf{Data}} & \textbf{Micro} & \textbf{Macro} & \textbf{Micro} & \textbf{Macro} & \multirow{-2}{*}{\textbf{Success}} \\ 

\midrule
Easy-1K  & 92.8 & 79.7 & 62.8 & 46.9 & 45.3 \\
Medium-1K & 94.9 & 69.2 & 66.2 & 49.0 & 41.1 \\ 
Hard-1K & 90.0 & 48.4 & 65.5 & 47.2 & 25.9 \\ 
\rowcolor{blue!10} \textbf{Mixed-1K} & \textbf{95.3} & \textbf{83.0} & \textbf{67.6} & \textbf{51.0} & \textbf{49.9} \\ 
\bottomrule
\end{tabular}
}
\end{minipage}
\end{figure}

\textbf{Motivation:} While SFT typically benefits from massive data volumes, the optimal data strategy for RL in complex agentic tasks remains underexplored. We investigate RL data composition across two orthogonal dimensions: \textbf{\textit{quantity}} and \textbf{\textit{difficulty}}. For quantity, we ask whether RL exhibits a continuous scaling law or a saturation point where over-optimization degrades generalization. For difficulty, we examine how the mixture of task complexity influences the resulting planning capabilities. Building upon our findings in previous sections, we fix the base model at 3B and utilize the \textsc{Curriculum} reward, the optimal configuration for models of this scale, to strictly isolate these data variables. Detailed experimental setups, training dynamics analysis, and full OOD results are deferred to Appendix~\ref{appendix:data_composition}.

% \noindent \textbf{Takeaway 4: RL data scaling exhibits a sweet spot; over-scaling degrades OOD generalization.} 
\begin{takeawaybox}
\textbf{Takeaway 4: RL data scaling exhibits a sweet spot; over-scaling degrades OOD generalization.}    
\end{takeawaybox}
As illustrated in \textbf{Figure~\ref{fig:data_quantity_test}}, increasing the training data from 100 to 1K prompts yields a rapid improvement in the in-domain success rate, rising from 37.5\% to 49.9\%. Concurrently, the average OOD score, shown in \textbf{Table~\ref{tab:data_ood}}, reaches its peak at 35.0\%. However, scaling further to 2K prompts causes a clear divergence. While the in-domain success rate marginally increases to 50.8\%, OOD generalization drops significantly to 32.2\%. This indicates that RL requires a modestly sized, high-quality data subset to effectively activate reasoning capabilities. Exceeding this sweet spot causes the model to over-optimize for the specific training distribution, sacrificing broader transferability for negligible in-domain gains.

% \noindent \textbf{Takeaway 5: Balanced data difficulty prevents reward sparsity and enables the mastery of complex constraints.} 

\begin{takeawaybox}
\textbf{Takeaway 5: Balanced data difficulty prevents reward sparsity and enables the mastery of complex constraints.} 
\end{takeawaybox}
\textbf{Table~\ref{tab:data_difficulty_indomain}} compares models trained on varying difficulty levels, discriminated by the number of constraints. For example, \textbf{easy} samples typically contain only a single budget limit, whereas \textbf{hard} samples introduce compounding requirements across transportation, meals, and accommodation. Training exclusively on easy data allows the model to grasp basic planning, achieving a high Commonsense Macro score of 79.7\%, but fails to teach complex constraint satisfaction. Conversely, training solely on hard data leads to a catastrophic performance collapse. The multifaceted constraints make successful trajectories exceedingly rare, exacerbating reward sparsity and preventing the model from learning even basic commonsense. The mixed configuration effectively resolves this dilemma. By blending difficulty levels, it provides enough simple tasks to maintain dense reward signals for commonsense learning, while incorporating sufficient complex tasks to teach advanced constraint satisfaction, ultimately achieving the highest overall success rate of 49.9\%.

\subsection{Algorithm Selection} 

\textbf{Motivation:} Recent advancements in agentic RL often introduce sophisticated sampling mechanisms to encourage exploration. To determine whether training long-horizon agents requires such algorithmic designs, we benchmark the standard GRPO against two representative variants: DAPO~\cite{yu2025dapo} and ARPO~\cite{dong2025arpo}. DAPO represents reward-guided trajectory filtering, e.g., discarding batches with zero variance in rewards, while ARPO represents adaptive rollout mechanisms that utilize entropy to dynamically branch trajectories. To provide appropriate reward signals across scales, we apply the \textsc{Macro} reward for the 1.5B and 3B models, and the \textsc{Sum} reward for the 7B model. To ensure a fair comparison, all algorithms share identical technical enhancements, with the key distinction lying solely in their exploration strategies. More implementation details and training dynamics are deferred to Appendix~\ref{appendix:algorithm}.

\begin{table*}[!t]
    \centering
    \caption{\textbf{Performance (\%) and training efficiency across different RL algorithms on the TravelPlanner test set.} Best results for each metric within the same model scale are in \textbf{bold}. GPU Hours are calculated based on 8$\times$A100-80G for 1.5B and 3B models and 16$\times$A100 for the 7B model.}
    \vspace*{-6pt}
    \resizebox{\linewidth}{!}{
    \begin{tabular}{ccccccccc}
       \toprule
      \rowcolor{COLOR_MEAN}  & & \multicolumn{2}{c}{\textbf{Commonsense}} & \multicolumn{2}{c}{\textbf{Hard Constraint}} & & & \\
      % \cmidrule(lr){3-4} \cmidrule(lr){5-6}
       \rowcolor{COLOR_MEAN} \multirow{-2}{*}{\textbf{Scale}} & \multirow{-2}{*}{\textbf{Algorithm}} &  \textbf{Micro} & \textbf{Macro} & \textbf{Micro} & \textbf{Macro} & \multirow{-2}{*}{\textbf{Success} \(\uparrow\)} & \multirow{-2}{*}{\textbf{Time / Step (min)} \(\downarrow\)} & \multirow{-2}{*}{\textbf{GPU Hours} \(\downarrow\)} \\ 
       \midrule
       \multirow{3}{*}{\textbf{1.5B}} 
       & GRPO & 93.4 & 68.4 & 47.9 & 31.6 & 30.1 & \cellcolor{orange!20}\textbf{8.0} & \cellcolor{orange!20}\textbf{164} \\ 
       & DAPO & \cellcolor{orange!10}\textbf{94.4} & \cellcolor{orange!10}\textbf{77.7} & 55.6 & 38.2 & 36.9 & 8.2 & 183 \\ 
       & ARPO & 93.1 & 76.4 & \cellcolor{orange!10}\textbf{58.1} & \cellcolor{orange!10}\textbf{39.4} & \cellcolor{orange!10}\textbf{37.5} & 9.5 & 195 \\ 
       \midrule 
       \multirow{3}{*}{\textbf{3B}} 
       & GRPO & 95.1 & 79.6 & \cellcolor{orange!10}\textbf{68.8} & \cellcolor{orange!10}\textbf{52.3} & \cellcolor{orange!10}\textbf{48.2} & \cellcolor{orange!20}\textbf{8.6} & \cellcolor{orange!20}\textbf{176} \\ 
       & DAPO & 93.6 & 78.6 & 66.1 & 48.7 & 45.6 & 8.9 & 184 \\ 
       & ARPO & \cellcolor{orange!10}\textbf{95.4} & \cellcolor{orange!10}\textbf{80.3} & 67.9 & 51.5 & 47.5 & 13.3 & 273 \\ 
       \midrule 
       \multirow{3}{*}{\textbf{7B}} 
       & GRPO & \cellcolor{orange!10}\textbf{96.9} & 87.4 & \cellcolor{orange!10}\textbf{78.7} & \cellcolor{orange!10}\textbf{66.5} & \cellcolor{orange!10}\textbf{62.8} & \cellcolor{orange!20}\textbf{9.0} & \cellcolor{orange!20}\textbf{368} \\ 
       & DAPO & 95.6 & \cellcolor{orange!10}\textbf{87.9} & 75.4 & 60.6 & 58.4 & 9.5 & 390 \\ 
       & ARPO & 96.7 & 86.8 & 76.6 & 61.1 & 58.3 & 13.3 & 547 \\ 
       \bottomrule
    \end{tabular}
    }
    \label{tab:algorithm_indomain_test}
    \vspace*{5pt}
\end{table*}

\noindent \textbf{Table~\ref{tab:algorithm_indomain_test}} presents a comprehensive comparison of both effectiveness and training efficiency across model scales, leading to the following insight.

% \noindent \textbf{Takeaway 6: The necessity for sophisticated exploration is inversely correlated with model capability.} 
\begin{takeawaybox}
\textbf{Takeaway 6: The necessity for sophisticated exploration is inversely correlated with model capability.}   
\end{takeawaybox}
As shown in Table~\ref{tab:algorithm_indomain_test}, smaller models inherently struggle with sparse reward signals. At the 1.5B scale, algorithmic interventions like ARPO and DAPO significantly outperform GRPO, achieving success rates of 37.5\% and 36.9\%, respectively, compared to 30.1\%. This advantage is also visually evident in their training dynamics in Appendix \textbf{Figure~\ref{fig:algorithm_training_dynamics}}, where exploration-heavy algorithms demonstrate faster convergence and higher peak performance early in training. However, as model capacity increases to 3B, this performance gap closes. Strikingly, at the 7B scale, GRPO achieves the highest success rate of 62.8\%, outperforming both DAPO and ARPO. Furthermore, sophisticated exploration introduces extra computational overhead. For instance, ARPO's adaptive rollout requires calculating the entropy difference at each generation step to determine whether to branch new trajectories. While intuitive, this mechanism increases the time cost per step. Therefore, these trends highlight a practical paradigm shift for scaling agentic RL: rather than expending computational resources and engineering effort on complex heuristic algorithms, we can leverage the inherent exploration capabilities of stronger base models, relying on the raw efficiency of the standard GRPO algorithm.

\subsection{Environmental Stability}

\begin{figure}[!t]
    \centering
    \begin{subfigure}[b]{0.45\textwidth}
        \centering
        \includegraphics[width=\textwidth]{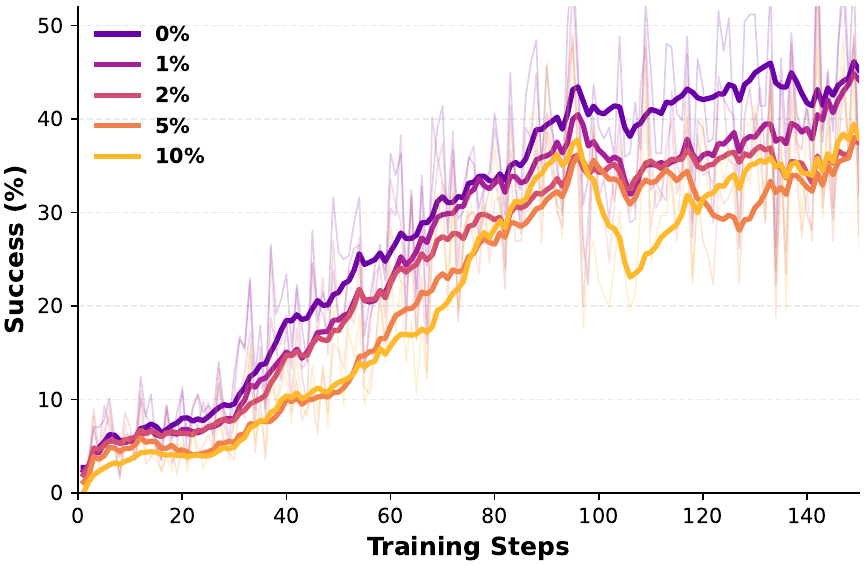}
        \caption{Training dynamics}
        \label{fig:env_stability_training}
    \end{subfigure}
    \begin{subfigure}[b]{0.45\textwidth}
        \centering
        \includegraphics[width=\textwidth]{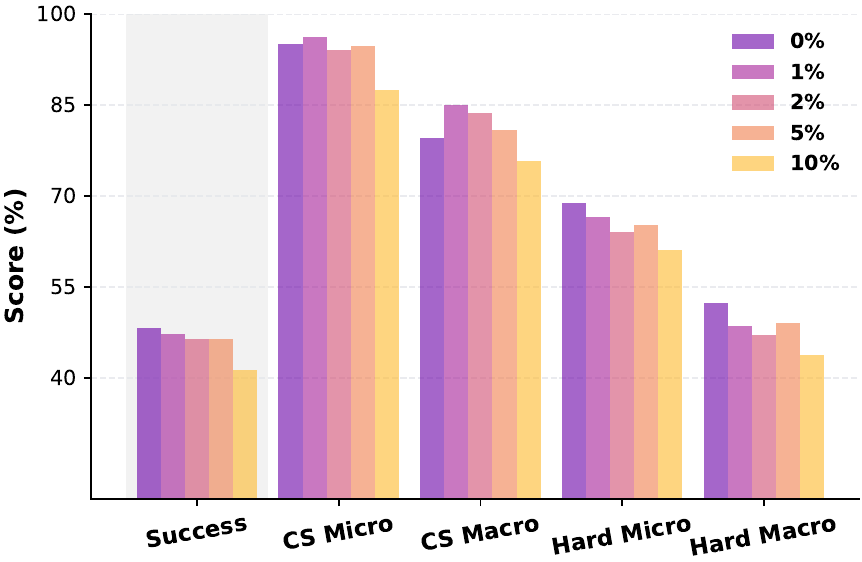}
        \caption{Test performance}
        \label{fig:env_stability_test}
    \end{subfigure}
    \vspace*{-5pt}
    \caption{\textbf{Training dynamics and test performance of the 3B model under varying environmental failure rates. } The percentages denote \textbf{the injected tool execution error rates} during RL training.
    \textbf{(a) Training dynamics:} High environmental instability, e.g., 10\% error rate, leads to slower convergence and increased variance. \textbf{(b) Test performance:} The agent maintains robust performance under mild noise but experiences a clear degradation across all metrics when the failure rate reaches 10\%.}
    \vspace*{-10pt}
\end{figure}

\noindent \textbf{Motivation:} Multi-turn tool calling naturally assumes a reliable execution environment, yet real-world APIs frequently suffer from unexpected errors. This raises a critical question: how robust is the RL training process when the environment itself is inherently unstable? To investigate this, we intentionally inject random tool execution failures during training. Specifically, we introduce a global error message ``\toolresponse{\textcolor{sftyellow}{\texttt{\textbf{Error: Current tool \{tool\_name\} is not available.}}}}'' into the observation space, which is triggered with a random probability during any tool execution. We vary the failure injection probability from 0\% to 10\% while keeping the reward design, i.e., \textsc{Macro}, and all other hyperparameters consistent. Finally, we evaluate the resulting models on a clean, failure-free test environment to isolate the impact of training instability on the learned policy. The training dynamics and the corresponding test performance are presented in \textbf{Figure~\ref{fig:env_stability_training}} and \textbf{Figure~\ref{fig:env_stability_test}}, respectively, leading to the following insight.

% \noindent \textbf{Takeaway 7: Agents exhibit resilience to mild environmental noise, but their performance experiences noticeable degradation under high instability.} 
\begin{takeawaybox}
\textbf{Takeaway 7: Agents exhibit resilience to mild environmental noise, but their performance experiences noticeable degradation under high instability.} 
\end{takeawaybox}
Initially, the 3B model maintains relatively stable test performance when the training failure rate remains at or below 5\%, as shown in Figure~\ref{fig:env_stability_test}. The final success rate fluctuates only slightly, indicating that the RL-trained agent can tolerate occasional environmental anomalies while still learning effective tool-use policies. However, when the failure rate reaches 10\%, the environmental instability introduces noticeable challenges. As illustrated in Figure~\ref{fig:env_stability_training}, this setting causes a clear drop in training stability and convergence speed. Consequently, the final test success rate drops significantly, accompanied by declines across all metrics. Ultimately, high environmental instability hinders the agent from completing successful trajectories during exploration, thereby starving the model of reliable reward signals necessary to optimize for complex constraints.

\section{Conclusion}
In this paper, we present a systematic empirical study to demystify the RL design space for long-horizon, tool-using agents. Using TravelPlanner as a testbed, we implement the flexible STAR pipeline to decompose the post-training process along 5 axes: reward shaping, model scaling, data composition, algorithm selection, and environmental stability. Our extensive empirical study yields a practical, scale-aware recipe: smaller models require staged curriculum rewards and sophisticated exploration, whereas larger models learn efficiently with simpler dense rewards and standard GRPO. Furthermore, we identified a data sweet spot in quantity and difficulty that balances in-domain mastery and OOD generalization. Applying this recipe, STAR achieves a new SOTA on TravelPlanner with much smaller models, significantly outperforming leading proprietary LLMs.

\noindent Our findings provide actionable guidelines for training long-horizon agents and highlight open challenges, e.g., how multiple factors simultaneously guide policy learning. We hope STAR serves as a valuable toolkit to inspire future research.

\textbf{Limitations:} Despite our comprehensive empirical study, several limitations warrant future investigation. (1) \textbf{Simulated environment:} While TravelPlanner is a challenging testbed, it remains simulated. Evaluating agents in real-world scenarios exposes them to diverse, unpredictable dynamics, more authentically reflecting their capability to handle practical, open-ended tasks. (2) \textbf{Limited OOD evaluation:} Our OOD evaluation is currently restricted to the knowledge-intensive QA task. This limits our conclusions to single-domain generalization, leaving the agents' cross-domain robustness largely unexplored. (3) \textbf{Unknown behavior of much larger LLMs:} Due to computational constraints, our study focuses on models up to 7B parameters. How the scale-aware recipe translates to the behaviors of massive LLMs remains an open question. (4) \textbf{Task-specific decomposition and single-factor analysis:} To ensure controlled comparisons, we primarily isolated individual axes, leaving the complex, simultaneous interactions among all dimensions unexplored. Furthermore, our reward shaping relies on task-specific, trajectory-level formulations.  Incorporating task-agnostic designs or fine-grained step-level bonuses could better alleviate sparse reward challenges in long-horizon tasks.

\clearpage
\newpage

\small
\bibliographystyle{nips}
\bibliography{reference}

\begin{thebibliography}{61}
\providecommand{\natexlab}[1]{#1}
\providecommand{\url}[1]{\texttt{#1}}
\expandafter\ifx\csname urlstyle\endcsname\relax
  \providecommand{\doi}[1]{doi: #1}\else
  \providecommand{\doi}{doi: \begingroup \urlstyle{rm}\Url}\fi

\bibitem[AI(2025)]{kimi-researcher2025}
Moonshot AI.
\newblock Kimi-researcher: End-to-end rl training for emerging agentic capabilities, 2025.
\newblock URL \url{https://moonshotai.github.io/Kimi-Researcher/}.

\bibitem[Bui et~al.(2026)Bui, Li, and Liu]{bui2026himaptravel}
The~Viet Bui, Wenjun Li, and Yong Liu.
\newblock Himap-travel: Hierarchical multi-agent planning for long-horizon constrained travel, 2026.
\newblock URL \url{https://arxiv.org/abs/2603.04750}.

\bibitem[Cheng et~al.(2026)Cheng, Hu, Zhang, Xu, Pan, Li, and Liu]{cheng2026travelbench}
Xiang Cheng, Yulan Hu, Xiangwen Zhang, Lu~Xu, Zheng Pan, Xin Li, and Yong Liu.
\newblock Travelbench: A broader real-world benchmark for multi-turn and tool-using travel planning, 2026.

\bibitem[Choi et~al.(2026)Choi, Yoon, Chen, Jha, and Pfister]{choi2026atlas}
Jihye Choi, Jinsung Yoon, Jiefeng Chen, Somesh Jha, and Tomas Pfister.
\newblock {ATLAS}: Constraints-aware multi-agent collaboration for real-world travel planning.
\newblock In \emph{The 14th International Conference on Learning Representations}, 2026.

\bibitem[DeepSeek-AI(2025)]{deepseekai2024deepseekv32exp}
DeepSeek-AI.
\newblock Deepseek-v3.2-exp: Boosting long-context efficiency with deepseek sparse attention, 2025.

\bibitem[Dong et~al.(2026)Dong, Mao, Ma, Bao, Chen, Wang, Chen, Du, Wang, Zhang, Zhou, Zhu, Wen, and Dou]{dong2025arpo}
Guanting Dong, Hangyu Mao, Kai Ma, Licheng Bao, Yifei Chen, Zhongyuan Wang, Zhongxia Chen, Jiazhen Du, Huiyang Wang, Fuzheng Zhang, Guorui Zhou, Yutao Zhu, Ji{-}Rong Wen, and Zhicheng Dou.
\newblock Agentic reinforced policy optimization.
\newblock In \emph{The 14th International Conference on Learning Representations (ICLR)}, 2026.

\bibitem[Feng et~al.(2025)Feng, Xue, Liu, and An]{feng2025groupingroup}
Lang Feng, Zhenghai Xue, Tingcong Liu, and Bo~An.
\newblock Group-in-group policy optimization for {LLM} agent training.
\newblock In \emph{The 39th Annual Conference on Neural Information Processing Systems}, 2025.

\bibitem[Ho et~al.(2020)Ho, Duong~Nguyen, Sugawara, and Aizawa]{2wiki}
Xanh Ho, Anh-Khoa Duong~Nguyen, Saku Sugawara, and Akiko Aizawa.
\newblock Constructing a multi-hop {QA} dataset for comprehensive evaluation of reasoning steps.
\newblock In Donia Scott, Nuria Bel, and Chengqing Zong (eds.), \emph{Proceedings of the 28th International Conference on Computational Linguistics}, pp.\  6609--6625, Barcelona, Spain (Online), December 2020. International Committee on Computational Linguistics.
\newblock \doi{10.18653/v1/2020.coling-main.580}.

\bibitem[Ji et~al.(2026)Ji, Ma, Wang, Chen, Chu, and Wu]{ji2025treegrpo}
Yuxiang Ji, Ziyu Ma, Yong Wang, Guanhua Chen, Xiangxiang Chu, and Liaoni Wu.
\newblock Tree search for llm agent reinforcement learning.
\newblock In \emph{The 14th International Conference on Learning Representations (ICLR)}, 2026.

\bibitem[Jimenez et~al.(2024)Jimenez, Yang, Wettig, Yao, Pei, Press, and Narasimhan]{jimenez2024swebench}
Carlos~E Jimenez, John Yang, Alexander Wettig, Shunyu Yao, Kexin Pei, Ofir Press, and Karthik~R Narasimhan.
\newblock {SWE}-bench: Can language models resolve real-world github issues?
\newblock In \emph{The 12th International Conference on Learning Representations (ICLR)}, 2024.

\bibitem[Jin et~al.(2025{\natexlab{a}})Jin, Yoon, Kargupta, Arik, and Han]{jin2025empirical}
Bowen Jin, Jinsung Yoon, Priyanka Kargupta, Sercan~O Arik, and Jiawei Han.
\newblock An empirical study on reinforcement learning for reasoning-search interleaved llm agents.
\newblock \emph{arXiv preprint arXiv:2505.15117}, 2025{\natexlab{a}}.

\bibitem[Jin et~al.(2025{\natexlab{b}})Jin, Zeng, Yue, Yoon, Arik, Wang, Zamani, and Han]{jin2025search}
Bowen Jin, Hansi Zeng, Zhenrui Yue, Jinsung Yoon, Sercan Arik, Dong Wang, Hamed Zamani, and Jiawei Han.
\newblock Search-r1: Training llms to reason and leverage search engines with reinforcement learning.
\newblock \emph{arXiv preprint arXiv:2503.09516}, 2025{\natexlab{b}}.

\bibitem[Joshi et~al.(2017)Joshi, Choi, Weld, and Zettlemoyer]{joshi2017triviaqa}
Mandar Joshi, Eunsol Choi, Daniel~S. Weld, and Luke Zettlemoyer.
\newblock Triviaqa: A large scale distantly supervised challenge dataset for reading comprehension, 2017.

\bibitem[Kwiatkowski et~al.(2019)Kwiatkowski, Palomaki, Redfield, Collins, Parikh, Alberti, Epstein, Polosukhin, Devlin, Lee, Toutanova, Jones, Kelcey, Chang, Dai, Uszkoreit, Le, and Petrov]{NQ}
Tom Kwiatkowski, Jennimaria Palomaki, Olivia Redfield, Michael Collins, Ankur Parikh, Chris Alberti, Danielle Epstein, Illia Polosukhin, Jacob Devlin, Kenton Lee, Kristina Toutanova, Llion Jones, Matthew Kelcey, Ming-Wei Chang, Andrew~M. Dai, Jakob Uszkoreit, Quoc Le, and Slav Petrov.
\newblock Natural questions: A benchmark for question answering research.
\newblock \emph{Transactions of the Association for Computational Linguistics}, 7:\penalty0 452--466, 2019.
\newblock \doi{10.1162/tacl_a_00276}.

\bibitem[Li et~al.(2025)Li, Zhang, Yin, Zhang, Ou, Wu, Yin, Li, Tao, Wang, et~al.]{li2025websailor}
Kuan Li, Zhongwang Zhang, Huifeng Yin, Liwen Zhang, Litu Ou, Jialong Wu, Wenbiao Yin, Baixuan Li, Zhengwei Tao, Xinyu Wang, et~al.
\newblock Websailor: Navigating super-human reasoning for web agent.
\newblock \emph{arXiv preprint arXiv:2507.02592}, 2025.

\bibitem[Li et~al.(2026{\natexlab{a}})Li, Zhang, Yin, Ye, Zhao, Zhang, Ou, Zhang, Wu, Wu, Wang, Qiao, Zhang, Jiang, Xie, Huang, and Zhou]{li2025websailorv2}
Kuan Li, Zhongwang Zhang, Huifeng Yin, Rui Ye, Yida Zhao, Liwen Zhang, Litu Ou, Dingchu Zhang, Xixi Wu, Jialong Wu, Xinyu Wang, Zile Qiao, Zhen Zhang, Yong Jiang, Pengjun Xie, Fei Huang, and Jingren Zhou.
\newblock Websailor-v2: Bridging the chasm to proprietary agents via synthetic data and scalable reinforcement learning.
\newblock In \emph{The 14th International Conference on Learning Representations (ICLR)}, 2026{\natexlab{a}}.

\bibitem[Li et~al.(2026{\natexlab{b}})Li, Guan, Zhang, Huang, Zhou, Lai, Yan, Jiang, Xie, Huang, Zhang, and Zhou]{li2026webweaver}
Zijian Li, Xin Guan, Bo~Zhang, Shen Huang, Houquan Zhou, Shaopeng Lai, Ming Yan, Yong Jiang, Pengjun Xie, Fei Huang, Jun Zhang, and Jingren Zhou.
\newblock Webweaver: Structuring web-scale evidence with dynamic outlines for open-ended deep research.
\newblock In \emph{The 14th International Conference on Learning Representations}, 2026{\natexlab{b}}.

\bibitem[Liu et~al.(2026)Liu, Guan, Nie, Zhang, Hao, Chen, Celikyilmaz, Wang, and Zhang]{liu2026alignmenttax}
Zhihan Liu, Lin Guan, Yixin Nie, Kai Zhang, Zhuoqun Hao, Lin Chen, Asli Celikyilmaz, Zhaoran Wang, and Na~Zhang.
\newblock Paying less generalization tax: A cross-domain generalization study of rl training for llm agents, 2026.

\bibitem[Lu et~al.(2026)Lu, Wang, Zhang, Wu, Gan, Zhuang, Gu, and Lin]{lu2025envtune}
Siyuan Lu, Zechuan Wang, Hongxuan Zhang, Qintong Wu, Leilei Gan, Chenyi Zhuang, Jinjie Gu, and Tao Lin.
\newblock Don't just fine-tune the agent, tune the environment.
\newblock In \emph{The 14th International Conference on Learning Representations (ICLR)}, 2026.

\bibitem[Mallen et~al.(2023)Mallen, Asai, Zhong, Das, Khashabi, and Hajishirzi]{mallen2023popqa}
Alex Mallen, Akari Asai, Victor Zhong, Rajarshi Das, Daniel Khashabi, and Hannaneh Hajishirzi.
\newblock When not to trust language models: Investigating effectiveness of parametric and non-parametric memories, 2023.

\bibitem[Ning et~al.(2025)Ning, Liu, Wang, Chen, Li, Fang, Zheng, Tan, and Liu]{ning2025deeptravel}
Yansong Ning, Rui Liu, Jun Wang, Kai Chen, Wei Li, Jun Fang, Kan Zheng, Naiqiang Tan, and Hao Liu.
\newblock Deeptravel: An end-to-end agentic reinforcement learning framework for autonomous travel planning agents, 2025.

\bibitem[OpenAI(2024)]{openai2024gpt4ocard}
OpenAI.
\newblock Gpt-4o system card, 2024.

\bibitem[OpenAI(2025{\natexlab{a}})]{openai2025gptoss120bgptoss20bmodel}
OpenAI.
\newblock gpt-oss-120b \& gpt-oss-20b model card, 2025{\natexlab{a}}.

\bibitem[OpenAI(2025{\natexlab{b}})]{singh2025openaigpt5card}
OpenAI.
\newblock Openai gpt-5 system card, 2025{\natexlab{b}}.

\bibitem[Ouyang et~al.(2022)Ouyang, Wu, Jiang, Almeida, Wainwright, Mishkin, Zhang, Agarwal, Slama, Ray, Schulman, Hilton, Kelton, Miller, Simens, Askell, Welinder, Christiano, Leike, and Lowe]{Ouyang2022TrainingLM}
Long Ouyang, Jeff Wu, Xu~Jiang, Diogo Almeida, Carroll~L. Wainwright, Pamela Mishkin, Chong Zhang, Sandhini Agarwal, Katarina Slama, Alex Ray, John Schulman, Jacob Hilton, Fraser Kelton, Luke~E. Miller, Maddie Simens, Amanda Askell, Peter Welinder, Paul~Francis Christiano, Jan Leike, and Ryan~J. Lowe.
\newblock Training language models to follow instructions with human feedback.
\newblock \emph{ArXiv}, abs/2203.02155, 2022.

\bibitem[Press et~al.(2023)Press, Zhang, Min, Schmidt, Smith, and Lewis]{bamboogle}
Ofir Press, Muru Zhang, Sewon Min, Ludwig Schmidt, Noah Smith, and Mike Lewis.
\newblock Measuring and narrowing the compositionality gap in language models.
\newblock In Houda Bouamor, Juan Pino, and Kalika Bali (eds.), \emph{Findings of the Association for Computational Linguistics: EMNLP 2023}, pp.\  5687--5711, Singapore, December 2023. Association for Computational Linguistics.
\newblock \doi{10.18653/v1/2023.findings-emnlp.378}.

\bibitem[Qwen et~al.(2025)Qwen, :, Yang, Yang, Zhang, Hui, Zheng, Yu, Li, Liu, Huang, Wei, Lin, Yang, Tu, Zhang, Yang, Yang, Zhou, Lin, Dang, Lu, Bao, Yang, Yu, Li, Xue, Zhang, Zhu, Men, Lin, Li, Tang, Xia, Ren, Ren, Fan, Su, Zhang, Wan, Liu, Cui, Zhang, and Qiu]{qwen2025qwen25technicalreport}
Qwen, :, An~Yang, Baosong Yang, Beichen Zhang, Binyuan Hui, Bo~Zheng, Bowen Yu, Chengyuan Li, Dayiheng Liu, Fei Huang, Haoran Wei, Huan Lin, Jian Yang, Jianhong Tu, Jianwei Zhang, Jianxin Yang, Jiaxi Yang, Jingren Zhou, Junyang Lin, Kai Dang, Keming Lu, Keqin Bao, Kexin Yang, Le~Yu, Mei Li, Mingfeng Xue, Pei Zhang, Qin Zhu, Rui Men, Runji Lin, Tianhao Li, Tianyi Tang, Tingyu Xia, Xingzhang Ren, Xuancheng Ren, Yang Fan, Yang Su, Yichang Zhang, Yu~Wan, Yuqiong Liu, Zeyu Cui, Zhenru Zhang, and Zihan Qiu.
\newblock Qwen2.5 technical report, 2025.

\bibitem[Schick et~al.(2023)Schick, Dwivedi-Yu, Dessì, Raileanu, Lomeli, Zettlemoyer, Cancedda, and Scialom]{schick2023toolformer}
Timo Schick, Jane Dwivedi-Yu, Roberto Dessì, Roberta Raileanu, Maria Lomeli, Luke Zettlemoyer, Nicola Cancedda, and Thomas Scialom.
\newblock Toolformer: Language models can teach themselves to use tools.
\newblock In \emph{The 37th Conference on Neural Information Processing Systems (NeurIPS)}, 2023.

\bibitem[Seed(2025)]{seedv18}
Bytedance Seed.
\newblock Seed1.8 model card: Towards generalized real-world agency, 2025.
\newblock URL \url{https://github.com/ByteDance-Seed/Seed-1.8}.

\bibitem[Shao et~al.(2026)Shao, Zhang, Yang, Chen, Han, Jinghao, Wei, Cai, Dong, Guo, and Li]{shao2026chinatravel}
Jie-Jing Shao, Bo-Wen Zhang, Xiao-Wen Yang, Baizhi Chen, Siyu Han, Pang Jinghao, Wen-Da Wei, Guohao Cai, Zhenhua Dong, Lan-Zhe Guo, and Yu-Feng Li.
\newblock Chinatravel: An open-ended travel planning benchmark with compositional constraint validation for language agents.
\newblock In \emph{The 14th International Conference on Learning Representations}, 2026.

\bibitem[Shao et~al.(2024)Shao, Wang, Zhu, Xu, Song, Bi, Zhang, Zhang, Li, Wu, and Guo]{shao2024deepseekmath}
Zhihong Shao, Peiyi Wang, Qihao Zhu, Runxin Xu, Junxiao Song, Xiao Bi, Haowei Zhang, Mingchuan Zhang, Y.~K. Li, Y.~Wu, and Daya Guo.
\newblock Deepseekmath: Pushing the limits of mathematical reasoning in open language models, 2024.

\bibitem[Shen et~al.(2026)Shen, Huang, Wang, Tian, Guo, Zhang, Zhou, Hu, Li, Xu, Wang, Liu, Li, Yue, Hong, Liu, and Zeng]{shen2026tripbench}
Yuanzhe Shen, Zisu Huang, Zhengyuan Wang, Muzhao Tian, Zhengkang Guo, Chenyang Zhang, Shuaiyu Zhou, Zengjie Hu, Dailin Li, Jingwen Xu, Kaimin Wang, Wenhao Liu, Tianlong Li, Fengpeng Yue, Feng Hong, Cao Liu, and Ke~Zeng.
\newblock Trip-bench: A benchmark for long-horizon interactive agents in real-world scenarios, 2026.

\bibitem[Tan et~al.(2025)Tan, Luo, Cai, Venkat, Montgomery, Hao, Wu, Balyan, Roongta, Wang, Li, Popa, and Stoica]{rllm2025}
Sijun Tan, Michael Luo, Colin Cai, Tarun Venkat, Kyle Montgomery, Aaron Hao, Tianhao Wu, Arnav Balyan, Manan Roongta, Chenguang Wang, Li~Erran Li, Raluca~Ada Popa, and Ion Stoica.
\newblock rllm: A framework for post-training language agents, 2025.
\newblock Notion Blog.

\bibitem[Team(2025)]{geminiteam2025}
Gemini Team.
\newblock Gemini: A family of highly capable multimodal models, 2025.

\bibitem[Team(2026{\natexlab{a}})]{kimiteam2026kimik25}
Kimi Team.
\newblock Kimi k2.5: Visual agentic intelligence, 2026{\natexlab{a}}.

\bibitem[Team(2026{\natexlab{b}})]{qwen35blog}
Qwen Team.
\newblock Qwen3.5: Accelerating productivity with native multimodal agents, February 2026{\natexlab{b}}.
\newblock URL \url{https://qwen.ai/blog?id=qwen3.5}.

\bibitem[Team et~al.(2025)Team, Li, Zhang, Zhang, Huang, Li, Chen, Yin, Wu, Zhou, et~al.]{tongyidr}
Tongyi~DeepResearch Team, Baixuan Li, Bo~Zhang, Dingchu Zhang, Fei Huang, Guangyu Li, Guoxin Chen, Huifeng Yin, Jialong Wu, Jingren Zhou, et~al.
\newblock Tongyi deepresearch technical report.
\newblock \emph{arXiv preprint arXiv:2510.24701}, 2025.

\bibitem[Trivedi et~al.(2022)Trivedi, Balasubramanian, Khot, and Sabharwal]{musique}
Harsh Trivedi, Niranjan Balasubramanian, Tushar Khot, and Ashish Sabharwal.
\newblock {M}u{S}i{Q}ue: Multihop questions via single-hop question composition.
\newblock \emph{Transactions of the Association for Computational Linguistics}, 10:\penalty0 539--554, 2022.
\newblock \doi{10.1162/tacl_a_00475}.

\bibitem[Wang et~al.(2025)Wang, Zou, Song, Feng, Fang, Lu, Liu, Luo, Liang, Huang, Zhong, Ye, Qin, Xiong, Song, Wu, Li, Li, Dun, Liu, Zan, Leng, Wang, Yu, Chen, Guo, Su, Huang, Shen, Shi, Yan, Zhao, Liu, Ye, Zheng, Xin, Zhao, Heng, Huang, Wang, Qin, Lin, Wu, Chen, Wang, Zhong, Zhang, Li, Li, Zhao, Jiang, Wu, Zhou, Pang, Han, Liu, Ma, Liu, Cai, Fu, Liu, Wang, Zhang, Zhou, Li, Shi, Yang, Tang, Li, Han, Lu, Lin, Tong, Li, Zhang, Miao, Jiang, Li, Zhao, Li, Ma, Lin, Zhang, Yang, Guo, Zhu, Liu, Du, Cai, Li, Yuan, Han, Wang, Guo, Cheng, Ma, Xiao, Huang, Chen, Du, Chen, Wang, Li, Yang, Zeng, Jin, Li, Chen, Chen, Chen, Zhao, and Shi]{wang2025uitars2technicalreportadvancing}
Haoming Wang, Haoyang Zou, Huatong Song, Jiazhan Feng, Junjie Fang, Junting Lu, Longxiang Liu, Qinyu Luo, Shihao Liang, Shijue Huang, Wanjun Zhong, Yining Ye, Yujia Qin, Yuwen Xiong, Yuxin Song, Zhiyong Wu, Aoyan Li, Bo~Li, Chen Dun, Chong Liu, Daoguang Zan, Fuxing Leng, Hanbin Wang, Hao Yu, Haobin Chen, Hongyi Guo, Jing Su, Jingjia Huang, Kai Shen, Kaiyu Shi, Lin Yan, Peiyao Zhao, Pengfei Liu, Qinghao Ye, Renjie Zheng, Shulin Xin, Wayne~Xin Zhao, Wen Heng, Wenhao Huang, Wenqian Wang, Xiaobo Qin, Yi~Lin, Youbin Wu, Zehui Chen, Zihao Wang, Baoquan Zhong, Xinchun Zhang, Xujing Li, Yuanfan Li, Zhongkai Zhao, Chengquan Jiang, Faming Wu, Haotian Zhou, Jinlin Pang, Li~Han, Qi~Liu, Qianli Ma, Siyao Liu, Songhua Cai, Wenqi Fu, Xin Liu, Yaohui Wang, Zhi Zhang, Bo~Zhou, Guoliang Li, Jiajun Shi, Jiale Yang, Jie Tang, Li~Li, Qihua Han, Taoran Lu, Woyu Lin, Xiaokang Tong, Xinyao Li, Yichi Zhang, Yu~Miao, Zhengxuan Jiang, Zili Li, Ziyuan Zhao, Chenxin Li, Dehua Ma, Feng Lin, Ge~Zhang, Haihua Yang, Hangyu Guo, Hongda Zhu,
  Jiaheng Liu, Junda Du, Kai Cai, Kuanye Li, Lichen Yuan, Meilan Han, Minchao Wang, Shuyue Guo, Tianhao Cheng, Xiaobo Ma, Xiaojun Xiao, Xiaolong Huang, Xinjie Chen, Yidi Du, Yilin Chen, Yiwen Wang, Zhaojian Li, Zhenzhu Yang, Zhiyuan Zeng, Chaolin Jin, Chen Li, Hao Chen, Haoli Chen, Jian Chen, Qinghao Zhao, and Guang Shi.
\newblock Ui-tars-2 technical report: Advancing gui agent with multi-turn reinforcement learning, 2025.

\bibitem[Wang et~al.(2024)Wang, Yang, Huang, Jiao, Yang, Jiang, Majumder, and Wei]{wang2024e5}
Liang Wang, Nan Yang, Xiaolong Huang, Binxing Jiao, Linjun Yang, Daxin Jiang, Rangan Majumder, and Furu Wei.
\newblock Text embeddings by weakly-supervised contrastive pre-training, 2024.

\bibitem[Wang et~al.(2026{\natexlab{a}})Wang, Zhang, Wang, Shi, Li, Han, Tong, Deng, Sun, Taylor, Zhu, Cong, Sun, and Wang]{wang2026arlarena}
Xiaoxuan Wang, Han Zhang, Haixin Wang, Yidan Shi, Ruoyan Li, Kaiqiao Han, Chenyi Tong, Haoran Deng, Renliang Sun, Alexander Taylor, Yanqiao Zhu, Jason Cong, Yizhou Sun, and Wei Wang.
\newblock Arlarena: A unified framework for stable agentic reinforcement learning, 2026{\natexlab{a}}.

\bibitem[Wang et~al.(2026{\natexlab{b}})Wang, Xie, Shen, Wang, and Yang]{wang2026rlanything}
Yinjie Wang, Tianbao Xie, Ke~Shen, Mengdi Wang, and Ling Yang.
\newblock Rlanything: Forge environment, policy, and reward model in completely dynamic rl system.
\newblock \emph{arXiv preprint arXiv:2602.02488}, 2026{\natexlab{b}}.

\bibitem[Wu et~al.(2024)Wu, Shen, Shan, Song, Wang, Zhang, Feng, Cheng, Chen, Xiong, and Li]{wu2024graph}
Xixi Wu, Yifei Shen, Caihua Shan, Kaitao Song, Siwei Wang, Bohang Zhang, Jiarui Feng, Hong Cheng, Wei Chen, Yun Xiong, and Dongsheng Li.
\newblock Can graph learning improve planning in llm-based agents?
\newblock In \emph{The 38th Conference on Neural Information Processing Systems (NeurIPS)}, 2024.

\bibitem[Wu et~al.(2025)Wu, Li, Zhao, Zhang, Ou, Yin, Zhang, Jiang, Xie, Huang, Cheng, Wang, Cheng, and Zhou]{wu2025resumun}
Xixi Wu, Kuan Li, Yida Zhao, Liwen Zhang, Litu Ou, Huifeng Yin, Zhongwang Zhang, Yong Jiang, Pengjun Xie, Fei Huang, Minhao Cheng, Shuai Wang, Hong Cheng, and Jingren Zhou.
\newblock Resum: Unlocking long-horizon search intelligence via context summarization.
\newblock \emph{arXiv preprint arXiv:2509.13313}, 2025.

\bibitem[Xi et~al.(2026)Xi, Huang, Liao, Huang, Liu, Guo, yajie yang, Zheng, Ye, Zhang, Chen, He, Ding, Li, Chen, Du, Yao, Xu, Chen, Gui, Wu, Zhang, Huang, and Jiang]{xi2026agentgymrl}
Zhiheng Xi, Jixuan Huang, Chenyang Liao, Baodai Huang, Jiaqi Liu, Honglin Guo, yajie yang, Rui Zheng, Junjie Ye, Jiazheng Zhang, Wenxiang Chen, Wei He, Yiwen Ding, Guanyu Li, Zehui Chen, Zhengyin Du, Xuesong Yao, Yufei Xu, Jiecao Chen, Tao Gui, Zuxuan Wu, Qi~Zhang, Xuanjing Huang, and Yu-Gang Jiang.
\newblock Agentgym-{RL}: An open-source framework to train {LLM} agents for long-horizon decision making via multi-turn {RL}.
\newblock In \emph{The 14th International Conference on Learning Representations}, 2026.

\bibitem[Xie et~al.(2024{\natexlab{a}})Xie, Zhang, Chen, Zhu, Lou, Tian, Xiao, and Su]{xie2024travelplanner}
Jian Xie, Kai Zhang, Jiangjie Chen, Tinghui Zhu, Renze Lou, Yuandong Tian, Yanghua Xiao, and Yu~Su.
\newblock Travelplanner: A benchmark for real-world planning with language agents.
\newblock In \emph{Forty-first International Conference on Machine Learning}, 2024{\natexlab{a}}.

\bibitem[Xie et~al.(2024{\natexlab{b}})Xie, Zhang, Chen, Li, Zhao, Cao, Hua, Cheng, Shin, Lei, Liu, Xu, Zhou, Savarese, Xiong, Zhong, and Yu]{OSWorld}
Tianbao Xie, Danyang Zhang, Jixuan Chen, Xiaochuan Li, Siheng Zhao, Ruisheng Cao, Toh~Jing Hua, Zhoujun Cheng, Dongchan Shin, Fangyu Lei, Yitao Liu, Yiheng Xu, Shuyan Zhou, Silvio Savarese, Caiming Xiong, Victor Zhong, and Tao Yu.
\newblock Osworld: Benchmarking multimodal agents for open-ended tasks in real computer environments.
\newblock In \emph{The 38th Conference on Neural Information Processing Systems (NeurIPS)}, 2024{\natexlab{b}}.

\bibitem[Yang et~al.(2024)Yang, Jimenez, Wettig, Lieret, Yao, Narasimhan, and Press]{yang2024sweagent}
John Yang, Carlos~E Jimenez, Alexander Wettig, Kilian Lieret, Shunyu Yao, Karthik~R Narasimhan, and Ofir Press.
\newblock {SWE}-agent: Agent-computer interfaces enable automated software engineering.
\newblock In \emph{The 38th Conference on Neural Information Processing Systems (NeurIPS)}, 2024.

\bibitem[Yang et~al.(2025)Yang, Jimenez, Zhang, Lieret, Yang, Wu, Press, Muennighoff, Synnaeve, Narasimhan, Yang, Wang, and Press]{yang2024swebenchmultimodal}
John Yang, Carlos~E. Jimenez, Alex~L. Zhang, Kilian Lieret, Joyce Yang, Xindi Wu, Ori Press, Niklas Muennighoff, Gabriel Synnaeve, Karthik~R. Narasimhan, Diyi Yang, Sida~I. Wang, and Ofir Press.
\newblock {SWE}-bench multimodal: Do ai systems generalize to visual software domains?
\newblock In \emph{The 13th International Conference on Learning Representations (ICLR)}, 2025.

\bibitem[Yang et~al.(2018)Yang, Qi, Zhang, Bengio, Cohen, Salakhutdinov, and Manning]{yang-etal-2018-hotpotqa}
Zhilin Yang, Peng Qi, Saizheng Zhang, Yoshua Bengio, William Cohen, Ruslan Salakhutdinov, and Christopher~D. Manning.
\newblock {H}otpot{QA}: A dataset for diverse, explainable multi-hop question answering.
\newblock In Ellen Riloff, David Chiang, Julia Hockenmaier, and Jun{'}ichi Tsujii (eds.), \emph{Proceedings of the 2018 Conference on Empirical Methods in Natural Language Processing}, pp.\  2369--2380, Brussels, Belgium, October-November 2018. Association for Computational Linguistics.
\newblock \doi{10.18653/v1/D18-1259}.

\bibitem[Yao et~al.(2023)Yao, Zhao, Yu, Du, Shafran, Narasimhan, and Cao]{yao2023react}
Shunyu Yao, Jeffrey Zhao, Dian Yu, Nan Du, Izhak Shafran, Karthik Narasimhan, and Yuan Cao.
\newblock {ReAct}: Synergizing reasoning and acting in language models.
\newblock In \emph{The 11th International Conference on Learning Representations (ICLR)}, 2023.

\bibitem[Yeo et~al.(2025)Yeo, Tong, Niu, Neubig, and Yue]{yeotong2025longcot}
Edward Yeo, Yuxuan Tong, Morry Niu, Graham Neubig, and Xiang Yue.
\newblock Demystifying long chain-of-thought reasoning in llms, 2025.

\bibitem[Yu et~al.(2026)Yu, Chen, Feng, Chen, Dai, Yu, Zhang, Ma, Liu, Wang, and Zhou]{yu2026memagent}
Hongli Yu, Tinghong Chen, Jiangtao Feng, Jiangjie Chen, Weinan Dai, Qiying Yu, Ya-Qin Zhang, Wei-Ying Ma, Jingjing Liu, Mingxuan Wang, and Hao Zhou.
\newblock Memagent: Reshaping long-context {LLM} with multi-conv {RL}-based memory agent.
\newblock In \emph{The 14th International Conference on Learning Representations}, 2026.

\bibitem[Yu et~al.(2025{\natexlab{a}})Yu, Zhang, Zhu, Yuan, Zuo, YuYue, Dai, Fan, Liu, Liu, Liu, Liu, Lin, Lin, Ma, Sheng, Tong, Zhang, Zhang, Zhang, Zhang, Zhu, Zhu, Chen, Chen, Wang, Yu, Song, Wei, Zhou, Liu, Ma, Zhang, Yan, Wu, and Wang]{yu2025dapo}
Qiying Yu, Zheng Zhang, Ruofei Zhu, Yufeng Yuan, Xiaochen Zuo, YuYue, Weinan Dai, Tiantian Fan, Gaohong Liu, Juncai Liu, LingJun Liu, Xin Liu, Haibin Lin, Zhiqi Lin, Bole Ma, Guangming Sheng, Yuxuan Tong, Chi Zhang, Mofan Zhang, Ru~Zhang, Wang Zhang, Hang Zhu, Jinhua Zhu, Jiaze Chen, Jiangjie Chen, Chengyi Wang, Hongli Yu, Yuxuan Song, Xiangpeng Wei, Hao Zhou, Jingjing Liu, Wei-Ying Ma, Ya-Qin Zhang, Lin Yan, Yonghui Wu, and Mingxuan Wang.
\newblock {DAPO}: An open-source {LLM} reinforcement learning system at scale.
\newblock In \emph{The 39th Annual Conference on Neural Information Processing Systems}, 2025{\natexlab{a}}.

\bibitem[Yu et~al.(2025{\natexlab{b}})Yu, Yang, Zou, Yan, and Wang]{yu2025demystify}
Zhaochen Yu, Ling Yang, Jiaru Zou, Shuicheng Yan, and Mengdi Wang.
\newblock Demystifying reinforcement learning in agentic reasoning.
\newblock \emph{arXiv preprint arXiv:2510.11701}, 2025{\natexlab{b}}.

\bibitem[Zhang et~al.(2026{\natexlab{a}})Zhang, Sun, Song, Qi, and Zheng]{zhang2026vsearcher}
Ruiyang Zhang, Qianguo Sun, Chao Song, Yiyan Qi, and Zhedong Zheng.
\newblock Vsearcher: Long-horizon multimodal search agent via reinforcement learning, 2026{\natexlab{a}}.

\bibitem[Zhang et~al.(2026{\natexlab{b}})Zhang, Xiong, Chen, Jia, Huang, Xu, and Zhang]{zhang2026rapo}
Siwei Zhang, Yun Xiong, Xi~Chen, Zi'an Jia, Renhong Huang, Jiarong Xu, and Jiawei Zhang.
\newblock Rapo: Expanding exploration for llm agents via retrieval-augmented policy optimization, 2026{\natexlab{b}}.

\bibitem[Zhang et~al.(2026{\natexlab{c}})Zhang, Jiang, Li, Tu, Su, Deng, Guo, Lv, and Lin]{zhang2026deepplanning}
Yinger Zhang, Shutong Jiang, Renhao Li, Jianhong Tu, Yang Su, Lianghao Deng, Xudong Guo, Chenxu Lv, and Junyang Lin.
\newblock Deepplanning: Benchmarking long-horizon agentic planning with verifiable constraints, 2026{\natexlab{c}}.

\bibitem[Zhao et~al.(2026)Zhao, Li, Wu, Zhang, Zhang, Li, Song, Chen, Wang, Wang, Jiang, Tu, Xie, Huang, and Zhou]{zhao2026repurposing}
Yida Zhao, Kuan Li, Xixi Wu, Liwen Zhang, Ding-Chu Zhang, Baixuan Li, Maojia Song, Zhuo Chen, Chenxi Wang, Xinyu Wang, Yong Jiang, Kewei Tu, Pengjun Xie, Fei Huang, and Jingren Zhou.
\newblock Repurposing synthetic data for fine-grained search agent supervision.
\newblock In \emph{The 14th International Conference on Learning Representations}, 2026.

\bibitem[Zheng et~al.(2025)Zheng, Liu, Li, Chen, Yu, Gao, Dang, Liu, Men, Yang, Zhou, and Lin]{gspo}
Chujie Zheng, Shixuan Liu, Mingze Li, Xiong-Hui Chen, Bowen Yu, Chang Gao, Kai Dang, Yuqiong Liu, Rui Men, An~Yang, Jingren Zhou, and Junyang Lin.
\newblock Group sequence policy optimization.
\newblock \emph{arXiv preprint arXiv:2507.18071}, 2025.

\bibitem[Zhu et~al.(2025)Zhu, Jiang, Sang, Tang, Song, He, Jain, Wang, and Geramifard]{zhu2025plannerr1}
Siyu Zhu, Yanbin Jiang, Hejian Sang, Shao Tang, Qingquan Song, Biao He, Rohit Jain, Zhipeng Wang, and Alborz Geramifard.
\newblock Planner-r1: Reward shaping enables efficient agentic rl with smaller llms, 2025.

\end{thebibliography}

\clearpage
\newpage
\appendix

\tableofcontents
\normalsize

\clearpage
\newpage

\section{Related Works}

\subsection{LLM-based Agents}
LLMs have evolved from static text generators into autonomous agents capable of reasoning, acting, and interacting with dynamic environments~\cite{yao2023react, schick2023toolformer}. This evolution has catalyzed applications across diverse domains, spanning deep research agents navigating the open web for information~\cite{li2025websailor, li2026webweaver, wu2025resumun, tongyidr, zhang2026vsearcher}, software engineering agents debugging real-world codebases~\cite{jimenez2024swebench, yang2024swebenchmultimodal, yang2024sweagent}, and GUI agents serving as personal assistants~\cite{OSWorld, wang2025uitars2technicalreportadvancing}.

Among existing agentic environments, we select TravelPlanner~\cite{xie2024travelplanner} as our primary testbed. As a pioneering planning benchmark~\cite{zhang2026deepplanning, shen2026tripbench, shao2026chinatravel}, it offers critical advantages for our empirical study. First, it features an abundant \textbf{1K-instance test set} with \textbf{real-world complexity}, demanding multi-tool orchestration and constraint satisfaction. Its inherent \textbf{difficulty} is evident, as even advanced models like Kimi-K2.5~\cite{kimiteam2026kimik25} achieve a mere 11.8\% success rate. Crucially for RL scaling, TravelPlanner provides a local sandbox that enables \textbf{zero-cost, high-concurrency} execution. In contrast, the computational overhead of GUI and code agents, coupled with the prohibitive API costs of open-web tasks, creates severe bottlenecks for the extensive environmental exploration. Therefore, TravelPlanner emerges as an ideal, scalable environment for studying agentic RL.

\subsection{Agentic Reinforcement Learning}
While RL from human or AI feedback~\cite{Ouyang2022TrainingLM} has become the standard practice for LLM alignment, applying RL to autonomous agents introduces unique challenges, including complex environmental dynamics, long-horizon requirements, and sparse, delayed reward signals~\cite{xi2026agentgymrl}. To tackle these issues, GRPO~\cite{shao2024deepseekmath} and its variants~\cite{feng2025groupingroup, gspo, yu2025dapo, zhang2026rapo} have emerged as the dominant algorithms for agentic RL. Building upon this foundation, recent advancements can be categorized into three primary directions:
(1) \textbf{Paradigm Adaptation:} To handle dynamic environments and long-horizon tasks, researchers are moving beyond the ReAct framework~\cite{yao2023react} to propose novel agentic workflows~\cite{wu2025resumun, yu2026memagent}. This shift has driven the development of specialized RL algorithms tailored to adapt models to these new paradigms. 
(2) \textbf{Exploration-driven Sampling:} To navigate vast action spaces and encourage exploration in highly uncertain states, recent methods modify the sampling process. A representative approach is ARPO~\cite{dong2025arpo}, which adaptively forks new paths upon detecting high entropy during generation.
(3) \textbf{Reward Shaping for Dense Supervision:} To mitigate the sparsity of outcome-based rewards, intermediate signals are increasingly introduced. For example, E-GRPO~\cite{zhao2026repurposing} allocates partial bonuses for retrieving key entities within the agent's reasoning traces during deep research tasks. Concurrently, DeepTravel~\cite{ning2025deeptravel} employs an LLM-as-a-Judge to provide step-wise, action-level rewards, combining them with trajectory-level success to form a more comprehensive supervisory signal.

Motivated by these progresses, we regard algorithm selection as a core dimension of our design space to investigate whether such sophisticated algorithms are necessary for long-horizon tasks.

\subsection{Empirical Studies of Agentic RL}
Beyond algorithmic innovations, recent empirical studies seek practical recipes for training LLM agents. For instance, \citet{jin2025empirical} analyze reward shaping and model selection, but their scope is limited to short-horizon interactions within the QA domain. While \citet{yu2025demystify} explore data composition and policy optimizations, their findings remain confined to mathematical reasoning. Although other works extend these evaluations to diverse agentic environments, focusing on training stability~\cite{wang2026arlarena} or cross-domain generalization~\cite{liu2026alignmenttax}, they predominantly rely on simplified text-based simulators. Consequently, these testbeds often fail to capture the multifaceted constraints and extensive trajectories inherent in complex, long-horizon planning tasks.

In summary, the research community currently lacks a holistic understanding of the practical recipes required to train long-horizon agents capable of utilizing diverse tools. To bridge this gap, we implement STAR, a post-training pipeline designed with the flexibility to seamlessly integrate and evaluate varying data compositions, reward functions, algorithms, and environmental dynamics. Leveraging this framework, we explore a vastly expanded design space.

\subsection{Progress in Travel Planning Agents}

Recent years have witnessed a surge of interest in autonomous travel planning, an area that perfectly exemplifies the challenges of long-horizon reasoning, tool-use, and multifaceted constraint satisfaction. Existing literature in this domain can be broadly categorized into three directions: 

\noindent \textbf{Extended Benchmarks:} Following the pioneering work of TravelPlanner, several new benchmarks have been proposed to evaluate planning capabilities under different contexts~\cite{ning2025deeptravel, shen2026tripbench, shao2026chinatravel, cheng2026travelbench}. Most of these benchmarks were recently released and can be considered concurrent work. We select TravelPlanner as our primary testbed because it remains the most classic and widely recognized benchmark in this community. Furthermore, it features a robust evaluation set of 1,000 instances, whereas most concurrent benchmarks are either not yet publicly available or lack comparable scale.

\noindent \textbf{Task-specific and Multi-agent Workflows:} To tackle the notoriously low success rates on TravelPlanner, numerous studies have engineered sophisticated, task-specific workflows, such as the multi-agent collaboration framework ATLAS~\cite{choi2026atlas}. While these approaches can boost overall performance, their heavily heuristic designs are often overly specialized to the travel planning domain. In contrast, we focus on the fundamental and universally adopted \textbf{ReAct} paradigm, investigating how to intrinsically enhance a single agent's long-horizon planning capabilities through RL. 

\noindent \textbf{Agentic RL for Planning:} Concurrently, a few pioneering works have begun applying RL to travel planning~\cite{bui2026himaptravel, ning2025deeptravel, zhu2025plannerr1}. For example, the aforementioned DeepTravel utilizes LLM-as-a-Judge for step-wise rewards~\cite{ning2025deeptravel}, and Planner-R1 introduces a staged reward shaping strategy~\cite{zhu2025plannerr1}. While we benchmark against the SOTA Planner-R1 to demonstrate the superior performance achieved by our STAR framework and recipe, our core contribution fundamentally differs. Rather than proposing a single algorithmic variant, we provide a holistic, multi-dimensional empirical study, decomposing the design space across reward shaping, model scaling, data composition, algorithm selection, and environmental stability, to distill a comprehensive recipe for agentic RL.

\clearpage
\newpage

\section{Details of TravelPlanner}\label{appendix:travelplanner_detail}

In this section, we provide supplementary details regarding the TravelPlanner benchmark~\cite{xie2024travelplanner}. This includes the tool execution sandbox, query examples across different difficulty levels, concrete evaluation rules, and the formatting process of the final plan.

\subsection{Sandbox}

\begin{table*}[!h]
    \centering
    \caption{\textbf{Details of implemented tools within the TravelPlanner testbed.}}
    \vspace*{-5pt}
    \label{tab:travelplanner_sandbox}
    \resizebox{\linewidth}{!}{
    \begin{tabular}{cccc}
       \toprule
       \rowcolor{COLOR_MEAN} \textbf{Tool}  & \textbf{Arguments} & \textbf{Description} & \textbf{\# Data Entries}  \\
       \midrule
        \texttt{SearchCity} &  {state} & Find cities within a specific state & 64  \\
       \texttt{SearchFlight} & \small{departure, destination, date} & Search for flight information between cities on specific dates & 3,827,361 \\ 
       \texttt{GoogleDistance} & \small{departure, destination, mode} & Calculate distance, travel time, and cost between two cities & 17,603 \\ 
        \texttt{SearchRestaurant} & {city} & Find restaurant options in a specific city & 9,552 \\
        \texttt{SearchAttraction} & {city} & Find tourist attractions in a specific city & 5,303 \\ 
        \texttt{SearchAccommodation} & {city} & Find accommodation options in a specific city & 5,064 \\ 
       \bottomrule
    \end{tabular}
    }
\end{table*}

\textbf{Table~\ref{tab:travelplanner_sandbox}} outlines the tools supported within the TravelPlanner environment, which enable agents to gather information across various aspects of travel. These tools are implemented by querying a local database, ensuring low latency and zero execution cost. To maintain alignment with the abundance of real-world information, all database entries were originally collected from real-world APIs by the creators of TravelPlanner, maximizing the reliability and authenticity of the sandbox.

\subsection{Instances}

\begin{table*}[!h]
    \centering
    \caption{\textbf{Examples of queries across different difficulty levels on the TravelPlanner validation set.} Different types of constraints are highlighted: \cbudget{budget}, \caccom{accommodation}, \ctrans{transportation}, \croute{routing}, and \ctime{duration}.}
    \label{tab:query_difficulty}
     \vspace*{-5pt}
    \small
    \begin{tabularx}{\textwidth}{cXX}
        \toprule
       \rowcolor{COLOR_MEAN} \textbf{Difficulty} & \textbf{Example Query} & \textbf{Description} \\
        \midrule
        
        % Easy Row
        \textbf{Easy} & 
        Please create a travel plan for me where I'll be departing from \croute{Washington} and heading to \croute{Myrtle Beach} for a \ctime{3-day} trip from \ctime{March 13th to March 15th, 2022}. Can you help me keep this journey within a budget of \cbudget{\$1,400}? & 
        \textbf{Single people \& Basic constraint:} Involves planning for 1 person. The only explicit constraint is a budget limit. \\
        \addlinespace[0.5em] 
        
        \textbf{Medium} & 
        We need to create a travel plan for 2 people departing from \croute{Lake Charles} and visiting \croute{2 cities in Texas}. The trip will last for \ctime{5 days}, starting from \ctime{March 4th to March 8th, 2022}. The budget is set at \cbudget{\$4,600}. In terms of accommodations, we prefer places where \caccom{parties are allowed}. & 
        \textbf{Specific constraints:} Requires planning for a group of people. Introduces specific accommodation preferences alongside the budget. \\
        \addlinespace[0.5em]
        
        \textbf{Hard} & 
        Could you create a \ctime{7-day} travel itinerary for 2 people, departing from \croute{Albuquerque} and visiting \croute{3 cities in Texas} from \ctime{March 8th to March 14th, 2022}? Our budget is set at \cbudget{\$5,000}. We require accommodations that \caccom{allow smoking} and are preferably \caccom{not shared rooms}. We would prefer to \ctrans{avoid any flights} for our transportation. & 
        \textbf{Multi-dimensional constraints:} Demands planning a long-horizon trip. Features multi-dimensional constraints like strict accommodation rules and transportation limitations. \\
        \bottomrule
    \end{tabularx}
\end{table*}

The TravelPlanner dataset is divided into three splits: a training set of 45 queries, a validation set of 180 queries, and a test set of 1,000 queries. To illustrate the nature of these queries, we present three examples of varying difficulty levels from the validation set in \textbf{Table~\ref{tab:query_difficulty}}. Difficulty is primarily determined by the number of constraints. \textbf{Easy} queries typically target a single traveler with only a total budget limit. \textbf{Medium} queries introduce additional constraints focusing on meals, transportation, or accommodation, often accompanied by an increase in the number of travelers, which complicates budget calculations. Finally, \textbf{Hard} queries impose even stricter constraints, presenting challenges for long-horizon planning that must satisfy multi-dimensional requirements simultaneously. Notably, both the difficulty levels and the trip durations are well-balanced across all dataset splits.

\begin{table*}[t]
    \centering
    \caption{\textbf{Official evaluation rules on TravelPlanner.}}
    \label{tab:evaluation_rules}
    \vspace*{-5pt}
    \small 
    \begin{tabularx}{\textwidth}{lX}
        \toprule
        \rowcolor{COLOR_MEAN} \multicolumn{2}{c}{\textit{\textbf{Commonsense}}} \\
        \midrule
        
        Within Sandbox & All information in the plan must be within the closed sandbox. Otherwise, it will be considered a hallucination. \\
        \addlinespace[0.3em]
        Complete Information & No key information should be left out of the plan, such as the lack of accommodation during travel. \\
        \addlinespace[0.3em]
        \croute{Within Current City} & All scheduled activities for the day must be located within that day's city(s). \\
        \addlinespace[0.3em]
        \croute{Reasonable City Route} & Changes in cities during the trip must be reasonable. \\
        \addlinespace[0.3em]
        \cmeal{Diverse Restaurants} & Restaurant choices should not be repeated throughout the trip. \\
        \addlinespace[0.3em]
        Diverse Attractions & Attraction choices should not be repeated throughout the trip. \\
        \addlinespace[0.3em]
        \ctrans{Non-conflict Transportation} & Transportation choices within the trip must be reasonable. For example, having both ``self-driving'' and ``flight'' would be considered a conflict. \\
        \addlinespace[0.3em]
        \caccom{Minimum Nights Stay} & The number of consecutive days spent in a specific accommodation during the trip must meet the corresponding required minimum number of nights' stay. \\
        
        \midrule
        \rowcolor{COLOR_MEAN} \multicolumn{2}{c}{\textit{\textbf{Hard Constraint}}} \\
        \midrule
        
        \cbudget{Budget} & The total budget of the trip. \\
        \addlinespace[0.3em]
        \caccom{Room Rule} & Room rules include ``No parties'', ``No smoking'', ``No children under 10'', ``No pets'', and ``No visitors''. \\
        \addlinespace[0.3em]
        \caccom{Room Type} & Room types include ``Entire Room'', ``Private Room'', ``Shared Room'', and ``No Shared Room''. \\
        \addlinespace[0.3em]
        \cmeal{Cuisine} & Cuisines include ``Chinese'', ``American'', ``Italian'', ``Mexican'', ``Indian'', ``Mediterranean'', and ``French''. \\
        \addlinespace[0.3em]

        \ctrans{Transportation} & Transportation options include ``No flight'' and ``No self-driving''. \\
        
        \bottomrule
    \end{tabularx}
    \setlength{\aboverulesep}{0.4ex}
    \setlength{\belowrulesep}{0.65ex}
\end{table*}

\subsection{Evaluation Rules}
Following the official evaluation protocol of TravelPlanner, we detail the specific rules for each evaluation dimension in Table~\ref{tab:evaluation_rules}, categorizing them into commonsense and hard constraints. For each dimension, the {micro score} is computed as the percentage of satisfied rules. If all rules within a dimension are satisfied, the {macro score} is set to 1. A generated trajectory is deemed a \textit{\textbf{Success}} only if both commonsense and hard constraints are fully satisfied. 

These constraints vividly illustrate the inherent complexity of travel planning. For instance, when selecting accommodations, the agent must satisfy not only commonsense rules, e.g., the minimum stay nights required by a specific hotel, but also the user's explicitly stated preferences, e.g., room type. Collectively, the agent must ensure that the entire itinerary fits within the overall budget, adheres to transportation rules, remains strictly grounded in the provided context to avoid hallucinations, and forms a logically consistent route.

\subsection{Planning Formatting}\label{appendix:plan_formatting}
Following the original evaluation protocol of TravelPlanner, the agent generates travel plans in a natural, creative language format, which we believe aligns with real-world user expectations and readability. However, for the purpose of automated evaluation, these natural language plans must be converted into a structured JSON format that details distinct daily elements, e.g., transportation, accommodation, and attractions. Rather than forcing the planning agent to directly output strict JSON, which could compromise its conversational naturalness and reasoning flow, we employ a dedicated formatting model to handle this conversion, ensuring that the structured JSON is used strictly for objective judgment.

While the official TravelPlanner benchmark utilizes GPT-4o~\cite{openai2024gpt4ocard} as the default formatting model, the massive scale of trajectory generation required for RL training and evaluation renders proprietary API calls prohibitively expensive. Therefore, we adopt DeepSeek-V3.2-Exp~\cite{deepseekai2024deepseekv32exp} as our consistent formatting model. To address potential concerns regarding evaluation fairness, we emphasize two points: 
First, \textbf{all checkpoints and comparative models in our study are evaluated using this exact same formatting pipeline}, ensuring a strictly fair comparison. 
Second, prior to large-scale deployment, we conducted a rigorous offline validation on a subset of sampled plans. DeepSeek-V3.2-Exp demonstrated exceptional instruction-following capabilities for structured formatting, achieving an alignment rate with human expert annotations that is comparable, if not marginally superior, to GPT-4o, particularly in avoiding minor information loss during complex natural-language-to-JSON conversions. This combination of cost-efficiency and high reliability makes it the optimal choice for our extensive RL evaluation pipeline.
\clearpage
\newpage

\section{Implementation Details}~\label{appendix:implementation} 
In this section, we detail our implementation, covering data synthesis, the default experimental setups for SFT and RL, and the inference protocol. Specific configurations for individual ablation studies are deferred to their respective analysis sections.

\subsection{Data Synthesis}
During the element sampling process, we balance the ratio of selecting easy, medium, and hard element combinations. For query generation, we randomly select with equal probability from strong open-weight models: GPT-OSS-120B~\cite{openai2025gptoss120bgptoss20bmodel} and DeepSeek-V3.2-Exp-671B~\cite{deepseekai2024deepseekv32exp}. We utilize different models for query generation to ensure a more diverse expression of travel intents. For the sampled elements, we first query the sandbox to verify whether the specific dates and constraints yield feasible solutions. Only the validated elements are fed into the LLM to be translated into natural language queries. For example, given the input: \texttt{\{ ``org'': ``Everett'', ``dest'': ``San Francisco'', ``days'': 3, ``date'': [``2022-06-16'', ``2022-06-17'', ``2022-06-18''], ``people\_number'': 1, ``hard\_constraint'': None, ``budget'': 1600 \}}, the DeepSeek-V3.2-Exp model translates it into the following query: ``Could you help me plan a 3-day trip from Everett to San Francisco, from June 16th to June 18th, 2022, with a budget of \$1,600 for one person?''

Through this iterative process, we curated over 10K queries with balanced difficulty levels for subsequent use. To ensure reliability, we randomly sampled 200 cases from this pool, on which the DeepSeek-V3.2-Exp-Thinking model achieved a success rate of 21.9\%, closely mirroring its 21.1\% performance on the 180-instance TravelPlanner validation set.

\subsection{Training Hyperparameters}

\textbf{SFT:} From the synthetic pool, we prompted DeepSeek-V3.2-Exp-Thinking on 5K cases, filtering for task success and strict format adherence. This yielded 1,198 valid trajectories with their statistics shown in Table~\ref{tab:sft_data}. Using the \texttt{LLaMA-Factory} framework\footnote{\url{https://github.com/hiyouga/LlamaFactory}}, we performed full-parameter fine-tuning with a training batch size of 32, a learning rate of \(5 \times 10^{-6}\), and a linear warmup ratio of 0.1. We trained the 3B and 7B models for 4 epochs, while the 1.5B model was trained for 6 epochs to compensate for its higher initial entropy.

\begin{table}[!t]
    \centering
    \caption{\textbf{Statistics of filtered trajectories on synthetic data.} ``Avg. Tokens of Planning'' denotes the average token cost of the final generated itinerary.}
    \label{tab:sft_data}
    \vspace*{2pt}
    \resizebox{0.85\linewidth}{!}{
    \begin{tabular}{c|cccc}
       \toprule
       \rowcolor{COLOR_MEAN}  \textbf{Type} & \textbf{\# Data Entries} & \textbf{Avg. Tool Calls} & \textbf{Avg. Tokens} & \textbf{Avg. Tokens of Planning} \\ 
       \midrule
        Easy & 627 (52.3\%) & 8.8 & 10.1K & 1204.7 \\ 
        Medium & 373 (31.1\%) & 8.4 & 9.8K & 1203.1 \\ 
        Hard & 198 (16.5\%) & 11.7 & 11.8K & 1343.1 \\ 
        All & 1,198 & 9.2 & 10.3K & 1227.1 \\
        \bottomrule
    \end{tabular}
    }
\end{table}

\noindent \textbf{RL:} Our default RL optimization utilizes the GRPO algorithm~\cite{shao2024deepseekmath}. The default training set comprises 1K synthetic samples (excluding SFT data) with an easy:medium:hard ratio of 4:3:3 to balance difficulty. We apply a maximum context length of 30K, a tool call budget of 60, a training batch size of 32, a group size of 8, a learning rate of \(2 \times 10^{-6}\) for 5 epochs, and a sampling \texttt{temperature} of 1.0. For reward computation, we first invoke DeepSeek-V3.2-Exp to parse the generated text plans into structured JSONs, enabling automated sub-metric evaluations, e.g., commonsense and hard constraints. The \textsc{\textbf{Curriculum}} reward is then implemented as: the \textsc{Sum} reward for the first two epochs, the \textsc{Macro} reward for the next two, and the binary \textsc{Success} reward for the final epoch, forming a gradual transition from dense to sparse. Model selection is based on the best performance on TravelPlanner's 180-instance validation set. Training for 7B models is distributed across two $8 \times$NVIDIA A100-80G nodes, while 1.5B and 3B models utilize a single node.

\subsection{Evaluation Setup}
Evaluation spans in-domain testing on the 1,000-instance TravelPlanner test set and OOD testing across 7 distinct knowledge-intensive QA benchmarks including Natural Questions (NQ)~\cite{NQ}, TriviaQA~\cite{joshi2017triviaqa}, PopQA~\cite{mallen2023popqa}, HotpotQA~\cite{yang-etal-2018-hotpotqa}, 2WikiMultiHopQA (2Wiki)~\cite{2wiki}, Musique~\cite{musique}, and Bamboogle~\cite{bamboogle}. 

For in-domain evaluation, inference on the TravelPlanner benchmark follows the standard ReAct paradigm. The agent performs iterative cycles of thinking, issuing tool calls, and receiving environmental feedback. The final natural language itinerary is parsed into a structured JSON format using DeepSeek-V3.2-Exp for automated evaluation via the official TravelPlanner Leaderboard\footnote{\url{https://huggingface.co/spaces/osunlp/TravelPlannerLeaderboard}}. To ensure fair comparisons, \textbf{all models are evaluated using this identical pipeline and codebase}.

\begin{itemize}
    \item \textbf{SOTA LLMs:} We evaluate recent strong models including DeepSeek-V3.2-Exp-Thinking-671B~\cite{deepseekai2024deepseekv32exp}, Gemini3-Pro~\cite{geminiteam2025}, Seed-1.8~\cite{seedv18}, Kimi-K2.5-1T~\cite{kimiteam2026kimik25}, Qwen3.5-122B-A10B, and Qwen3.5-397B-A17B~\cite{qwen35blog}. We set the maximum context length to 128K and the tool call budget to 100. The \texttt{temperature} is set to 1.0 for Kimi-K2.5-1T and 0.6 for others. We also include GPT-5~\cite{singh2025openaigpt5card} and Planner-R1-32B results from~\citet{zhu2025plannerr1}. 
    \item \textbf{Trained Checkpoints:} For our trained models and their Qwen2.5-Instruct base models, we consistently apply a \texttt{temperature} of 0.6, a \texttt{top\_p} of 0.95, a maximum context length of 32K, and a tool call budget of 60. 
\end{itemize}

For OOD evaluation, we utilize each dataset's official test split. The only available tool is a \texttt{Search} function that queries a local Wikipedia corpus. Following the practice of Search-R1~\cite{jin2025search}, we employ E5~\cite{wang2024e5} as the retriever. For each query, we retrieve the top-5 text snippets. Additionally, the maximum context length is set to 32K, with a tool call budget of 60. For all inference, trajectories exceeding the context limit, tool calling budget, or failing to output the required \answer{} format are terminated and marked as failures.

\clearpage
\newpage

\begin{figure*}[!t]
    \centering
    \includegraphics[width=\linewidth]{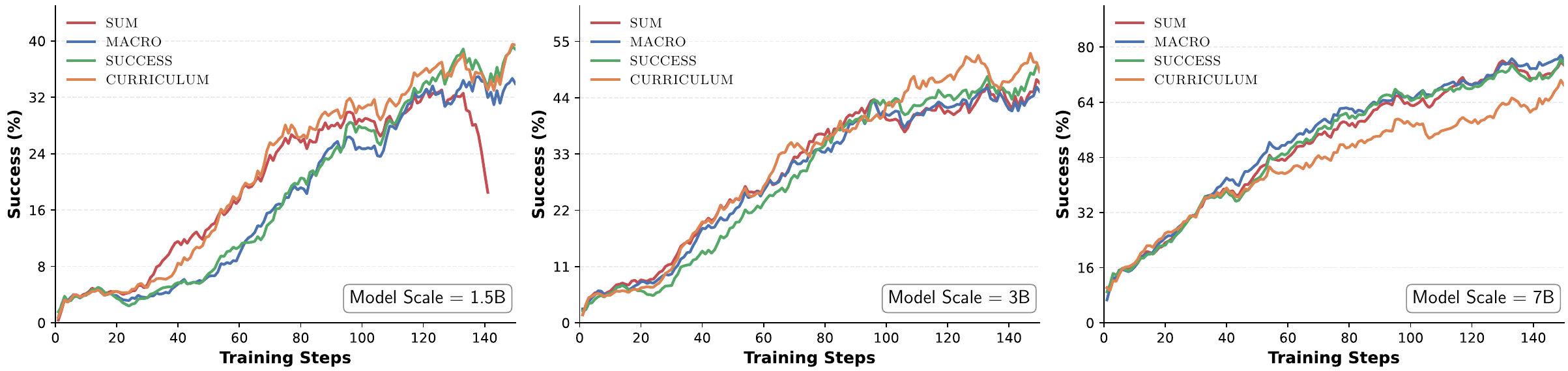}
    \caption{\textbf{Training dynamics (i.e., success rates on the training samples during each step) of different rewards.} The \textcolor{orange}{\textsc{\textbf{Curriculum}}} reward accelerates convergence and achieves higher peak performance for smaller models. Conversely, the dense rewards prove most effective for the stronger 7B model.}
    \label{fig:reward_training_dynamics}
\end{figure*}

\section{Supplementary Materials for Experiments}

\subsection{Reward Shaping}\label{appendix:reward}

\textbf{Setup:} To ensure a fair comparison, the only variable modified in this ablation is the reward computation strategy. All other settings strictly adhere to the default hyperparameter configuration. The specific epoch-level transitions for the \textsc{\textbf{Curriculum}} strategy have been detailed in the previous implementation section.

\noindent \textbf{Training Dynamics:} \textbf{Figure~\ref{fig:reward_training_dynamics}} illustrates the success rate curves over training steps, providing granular evidence for \textbf{Takeaway 1}, i.e., scale-dependent reward design. For the 1.5B and 3B models, the \textbf{\textcolor{orange}{\textsc{Curriculum}}} reward maintains the upper bound of performance, accelerating convergence. This shows that smaller models require progressive, staged learning signals. In contrast, the 7B model exhibits robust learning capacity, allowing the dense rewards to efficiently and stably optimize performance without requiring curriculum guidance.

\noindent \textbf{OOD Generalization:} \textbf{Table~\ref{tab:reward_ood}} details the OOD performance of different reward strategies on 7 knowledge-intensive QA benchmarks. To contextualize our results, we include three baselines: direct inference of the Base model, the SFT model, and Search-R1~\cite{jin2025search}, a strong baseline explicitly trained on its respective domain data. The results offer two critical insights: (1) \textbf{Transferability of complex tool-use:} Models trained exclusively on our TravelPlanner trajectories develop reasoning capabilities that transfer remarkably well to simpler domains. Across multiple scales, our RL variants achieve comparable or superior performance to the domain-specific Search-R1 baseline. (2) \textbf{Evidence of the alignment tax:} Table~\ref{tab:reward_ood} provides the empirical foundation for \textbf{Takeaway 2}. For the 7B model, while the dense \textsc{Sum} reward maximizes in-domain success (Table~\ref{tab:reward_indomain}), its OOD average accuracy drops to 36.7\%, falling significantly behind its SFT starting point (41.9\%). This demonstrates that overly dense, task-specific rewards cause the model to overfit to the TravelPlanner format, degrading general information-seeking abilities. The semi-sparse \textsc{Macro} reward mitigates this alignment tax, achieving the highest OOD performance (42.9\%) while remaining highly competitive in-domain.

\subsection{Model Scaling}\label{appendix:model_scale}

\begin{figure*}[!t]
    \centering
    \includegraphics[width=\linewidth]{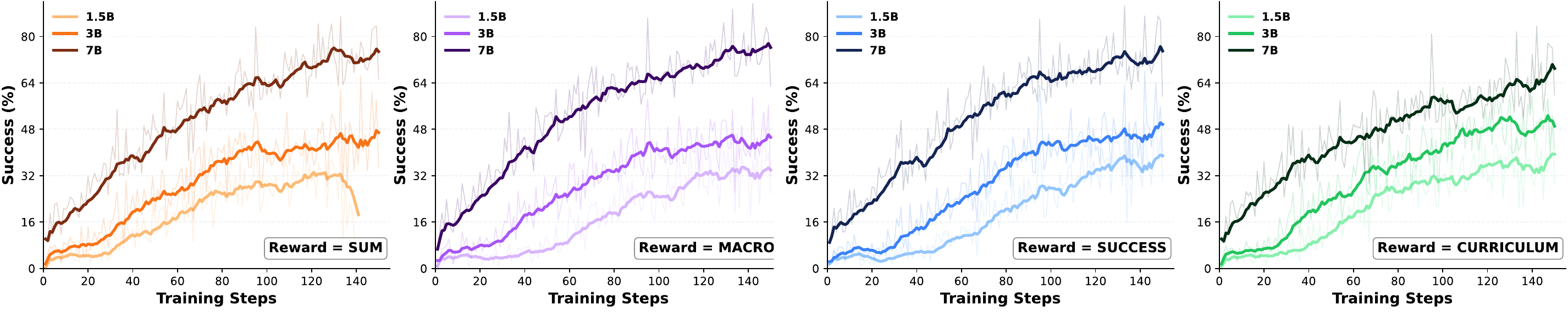}
    \caption{\textbf{Training dynamics across model scales.} For all reward types, increasing model capacity from 1.5B to 7B consistently leads to faster convergence, improved training stability, and higher performance asymptotes.}
    \label{fig:modelscale_training_dynamics}
\end{figure*}

\textbf{Training Dynamics:} To provide a more granular view of the scaling behaviors discussed in \textbf{Takeaway 3}, we present the training dynamics of the 1.5B, 3B, and 7B models across all four reward configurations in Figure~\ref{fig:modelscale_training_dynamics}. Scaling up the base model capacity yields three distinct advantages during the RL fine-tuning process: \textbf{accelerated convergence}, \textbf{enhanced training stability}, and \textbf{higher performance asymptotes}. Across all reward designs, the 7B models exhibit a significantly steeper initial learning curve compared to their 1.5B and 3B counterparts. For instance, under the sparse \textsc{Success} reward, the 1.5B model struggles to establish a clear upward sign in the early stages, whereas the 7B model rapidly acquires the necessary reasoning patterns, indicating that larger models possess stronger inherent priors to leverage RL signals efficiently. Beyond learning faster, larger models demonstrate superior robustness against the high variance typically associated with RL. The learning curves for the 1.5B model, exhibit fluctuations, whereas the 7B model maintains smoothness. Ultimately, larger models converge to higher final success rates, with the performance gaps widening over the course of training. This shows that the bottlenecks observed in smaller models are fundamental capacity limits. 

\subsection{Data Composition}\label{appendix:data_composition}

\textbf{Data Construction:} For all data composition experiments, we utilize the 3B model and apply the \textsc{Curriculum} reward to isolate the variables of data quantity and difficulty. 
\begin{itemize}
    \item \textit{\textbf{Quantity:}} The baseline is the 1K prompt set, which follows a 4:3:3 ratio of easy, medium, and hard samples. To create the 100, 200, and 500 subsets, we randomly subsample from this 1K baseline while preserving the 4:3:3 difficulty distribution. For the 2K setting, we augment the baseline with an additional 1K samples from the synthetic pool, adhering to the same difficulty ratio.
    \item \textit{\textbf{Difficulty:}} We independently construct three datasets, Easy-1K, Medium-1K, and Hard-1K, by sampling 1K instances comprised entirely of their respective difficulty levels from the remaining synthetic pool. These are compared against the default Mixed-1K baseline.
\end{itemize}

\begin{figure*}[!t]
    \centering
    \includegraphics[width=0.8\linewidth]{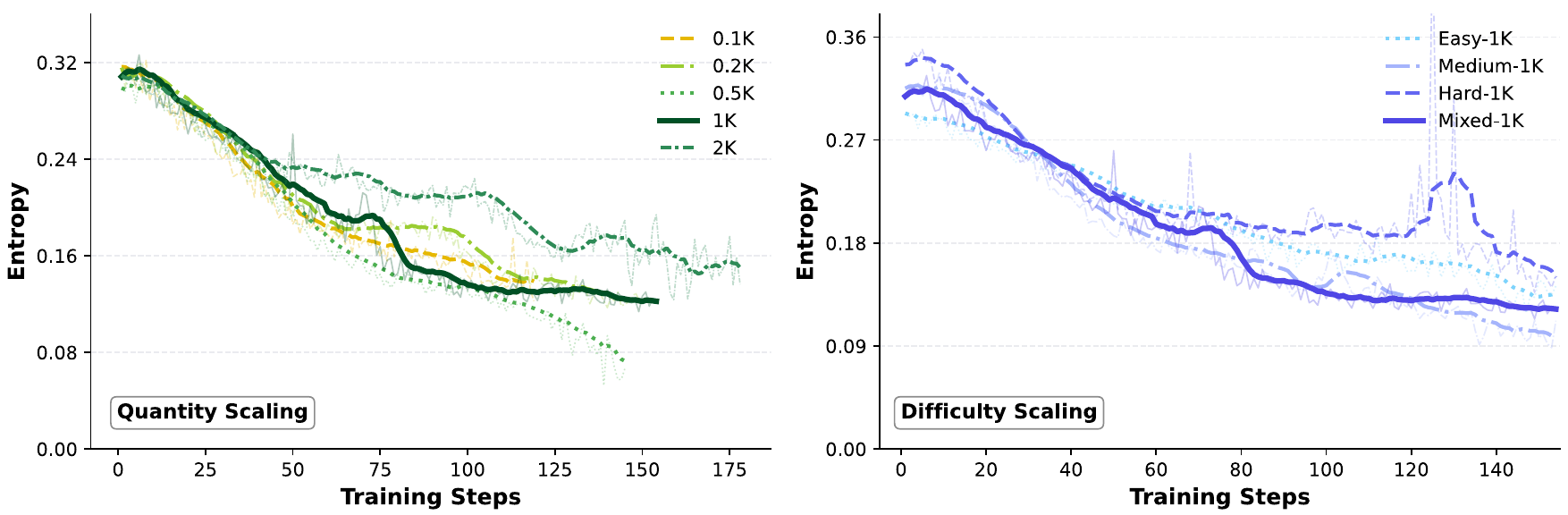}
    \caption{\textbf{Training dynamics for data quantity and difficulty experiments.} Configurations that yield the best generalization performance (1K and Mixed) are characterized by a stable, gradual decline in entropy. Suboptimal configurations suffer from anomalous training behaviors, e.g., premature entropy collapse and severe training instability.}
    \label{fig:data_scaling_entropy}
\end{figure*}

\noindent \textbf{Setup:} Varying the dataset size introduces a critical confounding factor: the number of gradient updates per epoch. To ensure a fair comparison and isolate the effect of the data itself, we adjust the total training epochs for different subset sizes. This guarantees that all models undergo \textbf{a total of approximately 155 gradient update steps}, which is equivalent to 5 epochs on the 1K dataset with a batch size of 32.

\noindent \textbf{Training Dynamics \& OOD Results:} Figure~\ref{fig:data_scaling_entropy} illustrates the training dynamics through action entropy, which reflects the model's exploration behavior. For data quantity, the optimal 1K configuration exhibits a smooth, gradual decline in entropy, whereas suboptimal settings like 0.5K display premature entropy collapse, indicating potential over-optimization. Regarding data difficulty, the Mixed-1K setting maintains a stable entropy descent. In contrast, the Hard-1K configuration suffers from severe training instability. Finally, \textbf{Table~\ref{tab:data_ood}} provides the comprehensive OOD results. Corroborating our in-domain findings, the 1K dataset size and the Mixed difficulty composition consistently yield the highest average OOD accuracy, confirming them as the sweet spots for robust generalization.

\begin{table*}[!t]
\centering
\caption{\textbf{Out-of-domain performance (in \%) on the knowledge-intensive QA benchmarks.} The best results of each dataset across data configurations are highlighted in \textbf{bold}.}
\vspace*{-5pt}
\label{tab:data_ood}
\resizebox{\linewidth}{!}{
\begin{tabular}{cccccccccc}
\toprule
\rowcolor{COLOR_MEAN} \textbf{Mode} & \textbf{Data} & \textbf{NQ} & \textbf{TriviaQA} & \textbf{PopQA} & \textbf{HotpotQA} & \textbf{2Wiki} & \textbf{Musique} & \textbf{Bamboogle} & \textbf{Avg.} \\
\midrule
\multicolumn{2}{c}{\textbf{SFT}} & 35.1 & 52.5 & 31.3 & 32.0 & 24.4 & 9.0 & 30.4 & 30.7 \\ \midrule

\multirow{5}{*}{\textbf{\textit{Quantity}}} & 0.1K & 39.2 & \textbf{57.4} & 35.0 & 38.5 & 26.3 & 9.8 & 30.4 & 33.8 \\ 
& 0.2K & 37.5 & 56.7 & 34.7 & 36.8 & \textbf{29.2} & 9.6 & 29.6 & 33.4 \\
& 0.5K & 39.9 & 57.3 & 34.0 & 37.4 & 25.2 & 11.8 & 29.6 & 33.6 \\ 
& 1K & \textbf{41.0} & 56.8 & \textbf{36.2} & \textbf{39.5} & 27.7 & \textbf{12.4} & \textbf{32.0} & \cellcolor{blue!10}\textbf{35.0} \\
& 2K & 38.5 & 56.7 & 34.4 & 34.3 & 22.4 & 9.6 & 29.6 & 32.2\\ 

\midrule 

\multirow{4}{*}{\textbf{\textit{Difficulty}}} & Easy-1K & 37.0 & 54.4 & 33.5 & 37.7 & \textbf{30.7} & 10.1 & 31.4 & 33.5 \\
& Medium-1K & 38.7 & 56.5 & 34.9 & 38.3 & 26.5 & 10.7 & 26.4 & 33.1\\ 
& Hard-1K & 38.1 & 55.1 & \textbf{36.8} & 39.0 & 29.9 & 11.1 & 28.0 & 34.0 \\
& Mixed-1K & \textbf{41.0} & \textbf{56.8} & 36.2 & \textbf{39.5} & 27.7 & \textbf{12.4} & \textbf{32.0} & \cellcolor{blue!10}\textbf{35.0} \\

\bottomrule
\end{tabular}
}
\end{table*}

\subsection{Algorithm Selection}\label{appendix:algorithm}

\begin{figure*}[!t]
    \centering
    \includegraphics[width=\linewidth]{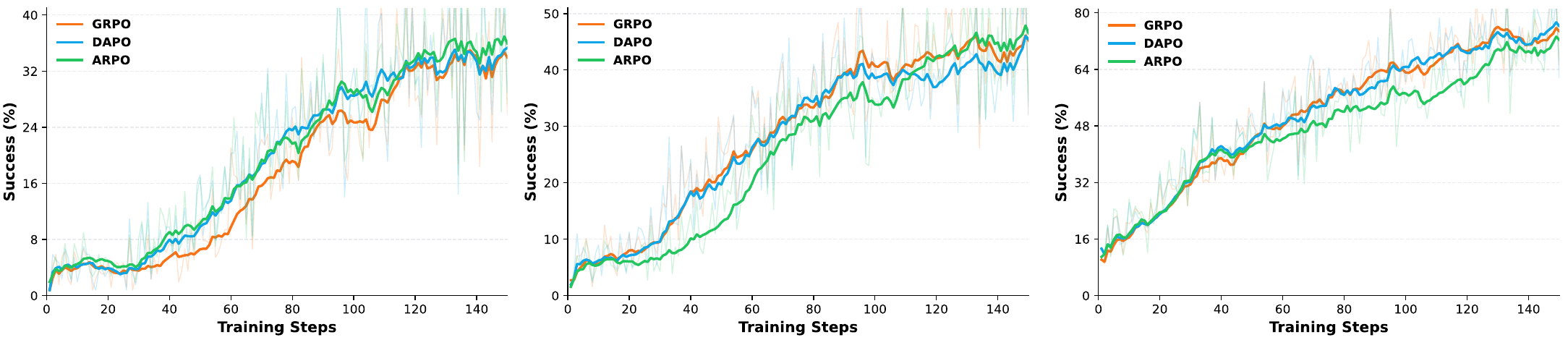}
    \caption{\textbf{Training dynamics of different RL algorithms.} The plots display the success rate curves during training for \textbf{1.5B (left)}, \textbf{3B (middle)}, and \textbf{7B (right)} models. Exploration-heavy algorithms like ARPO exhibit clear advantages at smaller scales, but this benefit diminishes as model capacity increases.}
    \label{fig:algorithm_training_dynamics}
\end{figure*}

\textbf{Setup:} To isolate the RL algorithm as the sole variable, all configurations utilize the default training data and hyperparameters. For the 1.5B and 3B models, although the \textsc{Curriculum} reward empirically yields the best performance, we apply the \textsc{Macro} reward to eliminate potential confounding factors introduced by reward switching. For the 7B model, we apply its optimal \textsc{Sum} reward. Across all three algorithms, technical enhancements designed to stabilize training, e.g., KL-free loss, remain consistent. Additionally, for DAPO, we maintain a batch size of 32 without further over-sampling.

\noindent \textbf{Training Dynamics:} Figure~\ref{fig:algorithm_training_dynamics} illustrates the training success rate curves for different RL algorithms. These dynamics further corroborate our findings in the main text: for smaller models, exploration-heavy algorithms demonstrate a clear advantage over standard GRPO. However, as model capacity scales to 3B and 7B, this algorithmic advantage diminishes. GRPO smoothly matches or even surpasses the complex variants, visually confirming that stronger base models rely less on heuristic exploration mechanisms.

\subsection{Recipe Summary}

Building on the preceding analyses, we synthesize the optimal training recipes tailored to each model scale. Regarding data configuration, we consistently apply the Mixed-1K dataset across all models, as it optimally balances in-domain proficiency with OOD generalization. For the 1.5B model, we adopt the ARPO algorithm to encourage exploration under high uncertainty. For the 3B model, we transition to the GRPO algorithm with the \textsc{Curriculum} reward. Finally, for the 7B model, we utilize GRPO combined with the dense \textsc{Sum} reward. The models trained with these specific configurations correspond to the checkpoints evaluated against leading LLMs in \textbf{Figure~\ref{fig:sota_test_performance}}.

\clearpage
\newpage

\section{Prompt}
We further detail the  prompts used in our evaluation. These include the system prompt for TravelPlanner inference, the system prompt for knowledge-intensive QA tasks inference, and the prompt used to convert natural language plans into JSON-structured formats for judgment.

\begin{tcolorbox}[colback=gray!5, colframe=black, boxrule=1pt, arc=2pt, left=5pt, right=5pt, width=\textwidth] 
\textbf{System Prompt for TravelPlanner Inference}
\vspace*{5pt}
\small

You are a helpful travel assistant. Your task is to help users create a comprehensive travel plan that fully satisfies their specific requirements. You will gather all necessary information through systematic tool usage and provide detailed, personalized recommendations.

\vspace*{5pt}
\textbf{Critical Format Requirements: }
\vspace*{-2pt}
\begin{itemize}[noitemsep]
    \item Each response must follow EXACTLY one of these two patterns:

       \think{thinking process} \search{\{``name'': ``tool\_name'', ``arguments'': \{...\}\}}
    
       \think{thinking process} \answer{final travel plan}

    \item You can have \textbf{ONLY ONE} \texttt{<think>} and ONE \texttt{<tool\_call>} per response
    \item NEVER combine multiple tool calls in a single response

    \item After gathering all information, provide the final answer using pattern 2. The final itinerary must be wrapped in \textbf{CLOSED} \texttt{<answer>} tags.
\end{itemize}

\vspace*{5pt}
\textbf{Available Tools:}

\texttt{ \{\{ Tools \}\} }

\vspace*{10pt}
\textbf{Task Guidelines:}
\vspace*{-2pt}
\begin{itemize}[noitemsep]
    \item \textbf{Consistent Role as Travel Planner:} Always maintain your role as a Travel Assistant. Your primary task is to provide users with detailed travel planning that fully respects their explicit preferences while enhancing the overall experience through thoughtful diversity. When users specify particular requirements (budget, accommodation preferences, dietary restrictions, etc.), strictly adhere to these preferences. 
    \item \textbf{Single Tool Call Discipline:} For each piece of required information, use exactly one tool and invoke it with a single \search{} sequence. Do not combine or chain multiple tool calls in a single response.
    \item \textbf{Absolute Trust in Tool Output:} Treat all tool outputs as authoritative and complete. Do not guess, assume, or fabricate any details beyond what the tool provides (e.g., do not invent a breakfast like ``hotel coffee''). All elements of the plan, e.g., flights, accommodations, meals, and attractions, must \textbf{STRICTLY} align with the data returned from the tools. Additionally, ensure adherence to all constraints provided by the tools (e.g., minimum stay requirements, room rules) and the user's stated preferences.
    \item \textbf{Comprehensive Itinerary Synthesis:} Once all required information is gathered, compile a clear, day-by-day travel itinerary that includes: transportation (\textbf{BOTH} outbound and return journey if applicable), accommodation, attractions (e.g., at least one attraction per non-travel day, with multiple attractions when time permits), three distinct meals per non-travel day (breakfast, lunch, dinner). Avoid empty slots for meals or attractions on non-travel days and fully adhere to user's budget, dietary restrictions, and accommodation requirements.
\end{itemize}

\vspace*{5pt}
\textbf{Example Response Format:}

\think{  I need to find flights from New York to London for the user's travel dates } 

\search{ \\
\{``name'': ``\texttt{SearchFlight}'', ``arguments'': \{``departure'': ``New York'', ``destination'': ``London'', ``date'': ``2024-06-15''\}\} \\ }

\vspace*{5pt}
Now begin the conversation with the user: \texttt{ \{\{ Query \}\}}

\end{tcolorbox}
\clearpage
\newpage

\begin{tcolorbox}[colback=gray!5, colframe=black, boxrule=1pt, arc=2pt, left=5pt, right=5pt, width=\textwidth] 
\textbf{Prompt for Knowledge-intensive QA Task Inference}
\vspace*{5pt}

\small

You are a specialized AI Search Assistant designed to answer questions by systematically retrieving and synthesizing information from a comprehensive knowledge base.

\vspace*{5pt}
\textbf{Critical Format Requirements: }
\vspace*{-2pt}
\begin{itemize}[noitemsep]
    \item Each response must follow EXACTLY one of these two patterns:

       \think{thinking process} \search{\{``name'': ``tool\_name'', ``arguments'': \{...\}\}}
    
       \think{thinking process} \answer{final answer}

    \item You can have \textbf{ONLY ONE} \texttt{<think>} and ONE \texttt{<tool\_call>} per response
    \item NEVER combine multiple tool calls in a single response

    \item After gathering all information, provide the final answer using pattern 2. The final itinerary must be wrapped in \textbf{CLOSED} \texttt{<answer>} tags.
\end{itemize}

\vspace*{5pt}
\textbf{Available Tools:}

\texttt{ \{\{ Tools \}\} }

\vspace*{5pt}
\textbf{Task Guidelines:}
\vspace*{-2pt}
\begin{itemize}[noitemsep]
    \item \textbf{Maintain Search Agent Identity:} You are a Search Assistant. Your core mission is to understand the user's question thoroughly, and then identify all information gaps that need to be filled. After systematically gathering relevant information using the Search tool, synthesize accurate, comprehensive answers from retrieved information.
    \item \textbf{Single Tool Call Discipline:} For each piece of required information, use exactly one tool and invoke it with a single \search{} sequence. Do not combine or chain multiple tool calls in a single response.
    \item \textbf{Absolute Trust in Tool Output:} Treat all tool outputs as authoritative and complete. Do not guess, assume, or fabricate any details beyond what the tool provides.
    \item \textbf{Synthesize Grounded Answers:} After gathering all required information, provide your thinking process in \think{} tags. Deliver a \textbf{CLEAR} and \textbf{CONCISE} final answer within \answer{} tags. Ensure your answer directly addresses the user's question.
\end{itemize}

\vspace*{5pt}
\textbf{Example Response Format:}

\think{I need to search for Leonardo DiCaprio's 1997 film. Let me retrieve information about his filmography from that year.} 

\search{ \\
\{``name'': ``\texttt{Search}'', ``arguments'': \{``query'': ``Leonardo DiCaprio 1997 film''\}\} \\ }

\vspace*{5pt}
Now begin the conversation with the user: \texttt{ \{\{ Query \}\}}

\end{tcolorbox}

\begin{tcolorbox}[colback=gray!5, colframe=black, boxrule=1pt, arc=2pt, left=5pt, right=5pt, width=\textwidth] 
\textbf{Prompt for Formatting Planning}
\vspace*{5pt}
\small

Please assist me in extracting valid information from a given natural language text and reconstructing it in JSON format, as demonstrated in the following example. 

\textbf{Important:}  You must output ONLY valid JSON format. Do not include any explanatory text, markdown code blocks, or additional formatting.

\vspace*{5pt}
\textbf{Guidelines:}
\vspace*{-2pt}
\begin{itemize}[noitemsep]
    \item Each item should include [`day', `current\_city', `transportation', `breakfast', `attraction', `lunch', `dinner', `accommodation']. Replace non-specific information like `eat at home/on the road' with `-'. Additionally, delete any `\$' symbols.
    \item Transportation: If transportation details indicate a journey from one city to another (e.g., from A to B), the `current\_city' should be updated to the destination city (in this case, B). If there's information about transportation, ensure that the `current\_city' aligns with the destination mentioned in the transportation details (i.e., the current city should follow the format `from A to B'). Also, ensure that all flight numbers and costs are followed by a colon (i.e., `Flight Number:' and `Cost:'), consistent with the provided example. For the `driving' or `drive' mode, unless otherwise specified, always use `Self-driving' consistently.
    \item Attraction: Use a `;' to separate different attractions, with each attraction strictly formatted as `Name, City'. 
    \item Day: start from 1.
\end{itemize}

\vspace*{5pt}
\textbf{Output Format:} 

Return ONLY the JSON array, nothing else.

\vspace*{5pt}
\textbf{Examples:}
\texttt{ \{\{ Examples \}\} }

\vspace*{5pt}
Now please help me extract valid information from the following text and reconstruct it in a strict JSON format. Your response must be \textbf{ONLY} the JSON array. Do not include any markdown code blocks, explanatory text, or other formatting. Just the raw JSON array starting with [ and ending with ].

\vspace*{5pt}
Original user query: \texttt{\{\{ Query \}\}}

Planning text:  \texttt{\{\{ Planning text \}\}}

\end{tcolorbox}

\clearpage
\newpage
\section{Cases}
In this section, we present the full trajectories on TravelPlanner's validation set generated by our STAR-trained agents, specifically focusing on the best checkpoints across different model scales.

We first showcase successful cases where the agent intelligently uses tools to gather relevant information, effectively balancing available choices for accommodations, meals, and transportation. This allows the agent to control the budget while strictly satisfying the user's requirements. Notably, in the first case, when the agent discovers that ground transportation is infeasible, it autonomously pivots to searching for flights. This demonstrates an emergent ability for \textit{\textbf{local error correction}} and persistent information gathering. In the second case, the agent exhibits robust long-horizon planning, seamlessly managing a multi-city itinerary under complex group and accommodation constraints.

To provide a comprehensive analysis and highlight avenues for future research, we also present failure cases that expose the inherent limitations of current LLM-based agents. First, we show a case illustrating that the agent can still struggle with \textbf{\textit{dense constraints}}. For example, the agent fails to fulfill the user's highly specific request for 4 distinct types of cuisines, indicating a degradation in instruction following when cognitive load is high.

More importantly, we highlight a critical structural limitation in complex planning: \textbf{\textit{the lack of global backtracking and re-planning}}. In our final failure case, the agent selects a destination city but subsequently discovers through tool calls that no accommodations in that city meet the user's strict requirements, e.g., a \caccom{``private room''} that allows \caccom{``parties''}. Instead of recognizing this as a dead end in the state space and backtracking to select a different city, the agent stubbornly proceeds. It compromises the hard constraints by selecting an invalid room type to justify its flawed trajectory.

\noindent \textbf{Future Direction:} This contrasting behavior, succeeding in local adaptation but failing in global re-planning, underscores a crucial gap in current agentic capabilities. It suggests that while our method significantly improves sequential decision-making, LLMs still struggle with long-horizon dependencies where upstream decisions must be revised based on downstream observations. Addressing this bottleneck remains a highly promising direction for future work in agentic planning and reasoning.

\begin{tcolorbox}[breakable, width=\linewidth, title=\textbf{Successful trajectory on TravelPlanner}]

\small

\textcolor{black}{\textbf{Question:} Please devise a travel plan that starts from \croute{St. Petersburg} and heads to \croute{Appleton}, taking place across \ctime{3 days} from \ctime{March 19th to March 21st, 2022}. This itinerary is for an individual, with a budget allocated at \cbudget{\$1,200}.}\\

\colorbox{blue!10}{%
    \parbox{\dimexpr\textwidth-2\fboxsep}{
    \centering
        \textbf{Turns 1 - 5: Transportation Planning \& Error Handling}
    }
}

\think{ \\ I need to start by checking the travel time and cost from St. Petersburg to Appleton on March 19th. I'll use GoogleDistanceMatrix to get the details.\\ }

\search{ \\ \{``name'': ``GoogleDistanceMatrix'', ``arguments'': \{``departure'': ``St. Petersburg'', ``destination'': ``Appleton'', ``mode'': ``driving''\}\} \\ }

\toolresponse{ \\ Sorry, we cannot find the distance between St. Petersburg and Appleton. \\}

\think{ \\ Since there's no direct route from St. Petersburg to Appleton via GoogleDistanceMatrix, I need to try another city in Wisconsin that might have better connectivity. I'll check Milwaukee first. \\ }

\search{ \\ \{``name'': ``GoogleDistanceMatrix'', ``arguments'': \{``departure'': ``St. Petersburg'', ``destination'': ``Milwaukee'', ``mode'': ``driving''\}\} \\ }

\toolresponse{ \\ No valid information found. Please rethink your distance search and try again. \\ }

\think{ \\ Since there's no direct route to Appleton or Milwaukee via GoogleDistanceMatrix, I need to try a different city in Wisconsin. Green Bay might have better connectivity. \\ }

\search{ \\ \{``name'': ``GoogleDistanceMatrix'', ``arguments'': \{``departure'': ``St. Petersburg'', ``destination'': ``Green Bay'', ``mode'': ``driving''\}\} \\ }

\toolresponse{ \\ No valid information found. Please rethink your distance search and try again. \\ }

\think{ \\ Since there's no direct route to Appleton or nearby cities via GoogleDistanceMatrix, I need to try a different approach. I'll \textbf{search for flights} from St. Petersburg to Appleton on March 19th. \\ }

\search{ \\ \{``name'': ``SearchFlight'', ``arguments'': \{``departure'': ``St. Petersburg'', ``destination'': ``Appleton'', ``date'': ``2022-03-19''\}\} \\ }

\toolresponse{ \\ A SearchFlight for St. Petersburg to Appleton on 2022-03-19 found the following 1 flights: \\
1. [F3574992] \$542. Departure: 22:51, Arrival: 00:42, Actual Elapsed Time: 2 hours 51 minutes, Distance: 1173.0 miles \\ }

\think{ \\ Now I need to search for the return flight from Appleton to St. Petersburg on March 21st. \\ }

\search{ \\ \{``name'': ``SearchFlight'', ``arguments'': \{``departure'': ``Appleton'', ``destination'': ``St. Petersburg'', ``date'': ``2022-03-21''\}\} \\ }

\toolresponse{ \\ A SearchFlight for Appleton to St. Petersburg on 2022-03-21 found the following 1 flights: \\
1. [F3578689] \$332. Departure: 07:19, Arrival: 11:01, Actual Elapsed Time: 2 hours 42 minutes, Distance: 1173.0 miles \\ }

\vspace*{10pt}

\colorbox{blue!10}{%
    \parbox{\dimexpr\textwidth-2\fboxsep}{%
    \centering
        \textbf{Turns 6 - 8: Accommodation, Attractions \& Dining Search}%
    }%
}

\think{\\ Now I need to search for accommodation in Appleton to complete the travel plan. \\}

\search{\\ \{``name'': ``SearchAccommodation'', ``arguments'': \{``city'': ``Appleton''\}\} \\}

\toolresponse{\\ A SearchAccommodation for Appleton found the following 14 accommodations: \\
1. [Nice bedroom in Brooklyn !] \$932.0. Room Type: Private room, House Rules: No pets, Minimum Nights: 4.0, Maximum Occupancy: 1, Review Rate Number: 5.0 \\ 
2. [Cozy room + bright apartment + rooftop lounge] \$666.0. Room Type: Private room, House Rules: No smoking, Minimum Nights: 1.0, Maximum Occupancy: 1, Review Rate Number: 4.0 \\
... \\}

\think{\\ Now I need to search for attractions in Appleton to continue building the itinerary. \\}

\search{\\ \{``name'': ``SearchAttraction'', ``arguments'': \{``city'': ``Appleton''\}\} \\}

\toolresponse{\\ A SearchAttraction for city Appleton found the following 20 attractions: \\
1. [The History Museum at the Castle](http://www.myhistorymuseum.org/) Address: 330 E College Ave, Appleton, WI 54911, USA. Contact: (920) 735-9370

2. [Hearthstone Historic House Museum](https://www.hearthstonemuseum.org/) Address: 625 W Prospect Ave, Appleton, WI 54911, USA. Contact: (920) 730-8204

... \\}

\think{\\ Now I need to search for restaurants in Appleton to complete the travel plan. \\}

\search{\\ \{``name'': ``SearchRestaurant'', ``arguments'': \{``city'': ``Appleton''\}\} \\}

\toolresponse{\\ A SearchRestaurant for city Appleton found the following 36 restaurants: \\
1. [Mikata Japanese Steakhouse] Avg. Cost \$72. Cuisines: Cafe, Pizza, Bakery, Italian. Rating: 3.6 \\
2. [Brick 29] Avg. Cost \$97. Cuisines: Desserts, Tea, BBQ, Cafe, American, Mediterranean. Rating: 4.4 \\
... \\}

\vspace*{10pt}
\colorbox{blue!10}{%
    \parbox{\dimexpr\textwidth-2\fboxsep}{%
    \centering
        \textbf{Turn 9: Final Itinerary Synthesis}%
    }%
}

\think{ \\ I have gathered all the necessary information from the tool calls. I need to compile the final travel plan with a day-by-day itinerary. \\} 

\answer{ \\ 
\textbf{Travel Plan: \croute{St. Petersburg} to \croute{Appleton} (\ctime{March 19-21, 2022})}  

\textbf{Budget:} \cbudget{\$1,200} | \textbf{Traveler:} Individual  

\textbf{Day 1: March 19 - Travel to Appleton}
\begin{itemize}[noitemsep]
    \item {Morning/Afternoon}: \ctrans{Flight} from \croute{St. Petersburg} to \croute{Appleton} [F3574992] | \cbudget{\$542} | Departure: 22:51 | Arrival: 00:42  
    \item {Evening}: Check into accommodation - \textbf{[\caccom{Stylish, convenient, renovated- 2 min to subway}]} | \cbudget{\$64/night} | Minimum Nights: 1 | Entire home/apt  
    % \item {Dinner}: At \textbf{\cmeal{Fat Lulu's}} | Avg Cost: \cbudget{\$23}  
    % \item {Attraction}: Visit \textbf{Plamann Park}  
\end{itemize}

\textbf{Day 2: March 20 - Explore Appleton}
\begin{itemize}[noitemsep]
    \item {Breakfast}: At \textbf{\cmeal{New Bakers Shoppee}} | Avg Cost: \cbudget{\$15}  
    \item {Morning Attraction}: Visit \textbf{Atlas Science Center}  
    \item {Lunch}: At \textbf{\cmeal{Side Wok}} | Avg Cost: \cbudget{\$15}  
    \item {Afternoon Attraction}: Visit \textbf{Trout Museum of Art}  
    \item {Dinner}: At \textbf{\cmeal{Mathew's Cafe}} | Avg Cost: \cbudget{\$13}   
\end{itemize}

\textbf{Day 3: March 21 - Return Travel}
\begin{itemize}[noitemsep]
    \item \textbf{Morning}: \ctrans{Flight} from \croute{Appleton} to \croute{St. Petersburg} [F3578689] | \cbudget{\$332} | Departure: 07:19 | Arrival: 11:01  
\end{itemize}

\textbf{Key Highlights} 
\begin{itemize}[noitemsep]
    \item \textbf{Total Budget}: \cbudget{\$1,200} 
    \item \textbf{Accommodation}: \caccom{Stylish, convenient} (2 nights) | \cbudget{\$128} total  
    \item \textbf{Flights}: \cbudget{\$874} total  
\end{itemize}

} \textcolor{red}{ \textbf{Satisfy both commonsense and hard constraints to achieve \textbf{\textit{Success}} $\checkmark$}}\\
\end{tcolorbox}

\clearpage
\newpage

\begin{tcolorbox}[breakable, title=\textbf{Successful trajectory on TravelPlanner}]

\small

\textcolor{black}{\textbf{Question:} Could you help develop a \ctime{week-long} travel itinerary suitable for a group of 6 people, departing from \croute{Baton Rouge} and planning to visit \croute{3 different cities in Texas}? The travel dates are set from \ctime{March 17th to March 23rd, 2022}. Our travel budget has been adjusted to \cbudget{\$14,600}. Bearing in mind that we have \caccom{children under ten years old}, our accommodations need to allow young children and we prefer to occupy \caccom{entire rooms}. We also prefer \ctrans{not to self-drive} during this trip.}\\

\colorbox{blue!10}{%
    \parbox{\dimexpr\textwidth-2\fboxsep}{
    \centering
        \textbf{Turns 1 - 5: Destination Selection \& Transportation Planning}
    }
}

\think{ \\ I need to start by \textbf{finding cities in Texas} to select three for the itinerary. I'll use the SearchCity tool to get a list of cities in Texas.\\ }

\search{ \\ \{``name'': ``SearchCity'', ``arguments'': \{``state'': ``Texas''\}\} \\ }

\toolresponse{ \\ A SearchCity for state Texas found the following 23 cities: Abilene, Amarillo, Harlingen, Lubbock, College Station, Corpus Christi, Wichita Falls, Waco, San Angelo, Houston, San Antonio, ... \\}

\think{ \\ I need to start by checking travel options from Baton Rouge to the first city. I'll \textbf{choose Houston as the first city} since it's a major city with good connectivity. I'll use GoogleDistanceMatrix to get travel details from Baton Rouge to Houston. \\ }

\search{ \\ \{``name'': ``GoogleDistanceMatrix'', ``arguments'': \{``departure'': ``Baton Rouge'', ``destination'': ``Houston'', ``mode'': ``taxi''\}\} \\ }

\toolresponse{ \\ A GoogleDistanceMatrix found the following information: taxi, from Baton Rouge to Houston, duration: 4 hours 0 mins, distance: 431 km, cost: \$431 \\ }

\think{ \\ Now I need to check travel from Houston to the second city. I'll \textbf{choose Austin as the second city} since it's a key city in Texas. I'll use GoogleDistanceMatrix to get travel details from Houston to Austin. \\}

\search{ \\ \{``name'': ``GoogleDistanceMatrix'', ``arguments'': \{``departure'': ``Houston'', ``destination'': ``Austin'', ``mode'': ``taxi''\}\} \\ }

\toolresponse{ \\ A GoogleDistanceMatrix found the following information:
taxi, from Houston to Austin, duration: 2 hours 31 mins, distance: 266 km, cost: \$266 \\ }

\textit{... [Model successfully repeats GoogleDistanceMatrix calls for  Austin $\rightarrow$ San Antonio, and San Antonio $\rightarrow$ Baton Rouge using Taxi]} \\

\colorbox{blue!10}{%
    \parbox{\dimexpr\textwidth-2\fboxsep}{%
    \centering
        \textbf{Turns 6 - 8: Accommodation Search}%
    }%
}

\think{\\ Now I need to search for accommodations in Houston. I'll use SearchAccommodation. \\}

\search{\\ \{``name'': ``SearchAccommodation'', ``arguments'': \{``city'': ``Houston''\}\} \\}

\toolresponse{\\ A SearchAccommodation for Houston found the following 10 accommodations: \\
1. [Sunny and Airy near Manhattan] \$499.0. Room Type: Entire home/apt, House Rules: No pets, Maximum Occupancy: 4... \\ 
2. [Bright room in 2BR apt in Nolita!] \$1024.0. Room Type: Private room, House Rules: No children under 10... \\
3. [Superhost 3 bedroom DISCOUNT] \$746.0. Room Type: Entire home/apt, House Rules: No pets, Minimum Nights: 1.0, Maximum Occupancy: ... \\ ... \\}

\textit{... [Model successfully repeats SearchAccommodation calls for Austin and San Antonio]} \\

\colorbox{blue!10}{%
    \parbox{\dimexpr\textwidth-2\fboxsep}{%
    \centering
        \textbf{Turns 9 - 14: Attractions \& Dining Search}%
    }%
}

\think{\\ Now I need to search for attractions in Houston to start building the itinerary. \\} 

\search{ \\
\{``name'': ``SearchAttraction'', ``arguments'': \{``city'': ``Houston''\}\} \\}

\toolresponse{\\ A SearchAttraction for city Houston found the following 20 attractions: \\ 
1. [Downtown Aquarium](http://www.aquariumrestaurants.com/downtownaquariumhouston/) Address: 410 Bagby St, Houston, TX 77002, USA. Contact: (713) 223-3474 

2. [Space Center Houston](https://www.spacecenter.org/) Address: 1601 E NASA Pkwy, Houston, TX 77058, USA. Contact: (281) 244-2100 

3. [Water Wall](https://uptown-houston.com/experience/parks/waterwall-park/) Address: 2800 Post Oak Blvd, Houston, TX 77056, USA. Contact: (713) 850-8841  \\ ... \\ }

\textit{... [Model continues to search for attractions in Austin and San Antonio, and then repeats SearchRestaurant for the three cities]} \\

\colorbox{blue!10}{%
    \parbox{\dimexpr\textwidth-2\fboxsep}{%
    \centering
        \textbf{Turn 15: Final Itinerary Synthesis}%
    }%
}

\think{ \\ I have gathered all the necessary information from the tool calls. I need to synthesize this into a comprehensive travel itinerary for a group of 6 people with a budget of \$14,600, visiting Houston, Austin, and San Antonio from March 17th to March 23rd, 2022. The itinerary will include travel details, accommodations, attractions, and meals. \\} 

\answer{ \\ 
\textbf{Week-long Travel Itinerary for 6 People}  

\textbf{Budget:} \cbudget{\$14,600} | \textbf{Dates:} \ctime{March 17–23, 2022} | \ctrans{No Self-Driving} | \caccom{Children Under 10} | \caccom{Entire Room Preferences}

\textbf{Day 1: March 17 – Travel to Houston}
\begin{itemize}[noitemsep]
    \item  Morning: \ctrans{Taxi} from \croute{Baton Rouge} to \croute{Houston} | Cost: \cbudget{\$431} | Duration: 4h  
    \item Afternoon: Check into [\caccom{Superhost 3 bedroom DISCOUNT}] | Entire home/apt | Cost: \cbudget{\$746/night}  
    \item Dinner: At \cmeal{Vinayaka Mylari} | Avg Cost: \cbudget{\$14/person} | Cuisines: Fast Food, French, Cafe, Italian
    \item Attraction: Visit \textbf{Hermann Park} | Contact: (713) 526-2183
\end{itemize}

\textbf{Day 2: March 18 – Explore Houston}
\begin{itemize}[noitemsep]
    \item Breakfast: At \cmeal{Super Bakery} | Avg Cost: \cbudget{\$30/person} | Cuisines: Seafood, Fast Food
    \item Morning Attraction: \textbf{Houston Museum of Natural Science} | Contact: (713) 639-4629
    \item Lunch: At \cmeal{Matchbox} | Avg Cost: \cbudget{\$33/person} | Cuisines: Tea, Italian, Bakery, Fast Food, Seafood
    \item Afternoon Attraction: \textbf{Space Center Houston} | Contact: (281) 244-2100
    \item Dinner: At \cmeal{Earthen Spices} | Avg Cost: \cbudget{\$28/person} | Cuisines: Tea, Mexican, Bakery, BBQ, Fast Food
\end{itemize}

\textbf{Day 3: March 19 – Travel to Austin}
\begin{itemize}[noitemsep]
    \item Morning: \ctrans{Taxi} from \croute{Houston} to \croute{Austin} | Cost: \cbudget{\$266} | Duration: 2h 31m  
    \item Afternoon: Check into [\caccom{Spacious Williamsburg 1 bedroom}] | Entire home/apt | Cost: \cbudget{\$455/night} 
    \item Dinner: At \cmeal{Moksha} | Avg Cost: \cbudget{\$12/person} | Cuisines: Cafe, Bakery, BBQ, Fast Food
    \item Attraction: Visit \textbf{Zilker Metropolitan Park} | Contact: (512) 974-6700
\end{itemize}

\textit{... [Days 4-6 omitted for brevity, successfully detailing \croute{Austin} and \croute{San Antonio} with \ctrans{Taxi} transfers, \caccom{Entire home/apt} stays, and 3 \cmeal{meals} per day]}

\vspace*{10pt}

\textbf{Day 7: March 23 – Return to Baton Rouge}
\begin{itemize}[noitemsep]
    \item Morning: \ctrans{Taxi} from \croute{San Antonio} to \croute{Baton Rouge} | Cost: \cbudget{\$746} | Duration: 6h 52m  
\end{itemize}

\textbf{Budget Summary} 

\begin{itemize}[noitemsep]
    \item Total Budget: \cbudget{\$14,600}  
    \item Estimated Costs: \cbudget{\$13,500} (within budget)  
    \item Remaining: \cbudget{\$1,100}  
\end{itemize}

\textbf{Key Highlights}

\begin{itemize}[noitemsep]
    \item Accommodations: \caccom{Entire rooms selected for children}.  
    \item Transport: \ctrans{Taxi} used for all segments  
    \item Attractions: Visited key sites in each city  
    \item Meals: Varied \cmeal{cuisines} covered  
\end{itemize}
} \textcolor{red}{ \textbf{Satisfy both commonsense and hard constraints to achieve \textbf{\textit{Success}} $\checkmark$}}\\
\end{tcolorbox}

\clearpage
\newpage

\begin{tcolorbox}[breakable, width=\linewidth, title=\textbf{Failure trajectory on TravelPlanner: Incomplete Constraint Satisfaction}]

\small

\textcolor{black}{\textbf{Question:} Can you please generate a 3-day travel plan for 2 people departing from \croute{New York} and traveling to \croute{Charleston} from \ctime{March 19th to March 21st, 2022}? Our budget is set at \cbudget{\$1,600}. We require accommodations that \caccom{allow smoking} and should \caccom{not be shared rooms}. We have a preference for \cmeal{French, Mediterranean, Mexican, and Chinese cuisines} during our trip.}\\

\colorbox{blue!10}{%
    \parbox{\dimexpr\textwidth-2\fboxsep}{
    \centering
        \textbf{Turns 1 - 2: Transportation Planning}
    }
}

\think{ \\ I need to find flight information from New York to Charleston on March 19th, 2022, as this is the departure date for the travel plan.\\ }

\search{ \\ \{``name'': ``SearchFlight'', ``arguments'': \{``departure'': ``New York'', ``destination'': ``Charleston'', ``date'': ``2022-03-19''\}\} \\ }

\toolresponse{ \\ A SearchFlight for New York to Charleston on 2022-03-19 found the following 8 flights: \\
... \\
8. [F4066693] \$137. Departure: 17:48, Arrival: 19:56, Actual Elapsed Time: 2 hours 8 minutes, Distance: 641.0 miles \\}

\think{ \\ I need to find return flight information from Charleston to New York on March 21st, 2022, as this is the return date for the travel plan. \\ }

\search{ \\ \{``name'': ``SearchFlight'', ``arguments'': \{``departure'': ``Charleston'', ``destination'': ``New York'', ``date'': ``2022-03-21''\}\} \\ }

\toolresponse{ \\ A SearchFlight for Charleston to New York on 2022-03-21 found the following 11 flights: \\
1. [F3635917] \$143. Departure: 11:01, Arrival: 12:58, Actual Elapsed Time: 1 hours 57 minutes, Distance: 636.0 miles \\
... \\ }

\vspace*{10pt}

\colorbox{blue!10}{%
    \parbox{\dimexpr\textwidth-2\fboxsep}{%
    \centering
        \textbf{Turn 3: Accommodation Search}%
    }%
}

\think{\\ I need to find accommodation options in Charleston that allow smoking and are not shared rooms, as per the user's requirements. \\}

\search{\\ \{``name'': ``SearchAccommodation'', ``arguments'': \{``city'': ``Charleston''\}\} \\}

\toolresponse{\\ A SearchAccommodation for Charleston found the following 20 accommodations: \\
1. [2 Bedroom Apartment in Prime Midtown West] \$1121.0. Room Type: Entire home/apt, House Rules: No smoking \& No children under 10... \\ 
2. [Bronx Roon] \$127.0. Room Type: Private room, House Rules: No visitors \& No children under 10, Minimum Nights: 2.0, Maximum Occupancy: 2, Review Rate Number: 2.0 \\
... \\}

\vspace*{10pt}

\colorbox{blue!10}{%
    \parbox{\dimexpr\textwidth-2\fboxsep}{%
    \centering
        \textbf{Turns 4 - 5: Attractions \& Dining Search}%
    }%
}

\think{\\ I need to find restaurant options in Charleston that match the user's preferred cuisines: French, Mediterranean, Mexican, and Chinese. \\}

\search{\\ \{``name'': ``SearchRestaurant'', ``arguments'': \{``city'': ``Charleston''\}\} \\}

\toolresponse{\\ A SearchRestaurant for city Charleston found the following 62 restaurants: \\
1. [\textbf{Filling Station}] Avg. Cost \$78. Cuisines: Desserts, Tea, Italian, \textbf{French}, Cafe. Rating: 4.1 \\
...\\
4. [\textbf{Makansutra Gluttons Bay}] Avg. Cost \$84. Cuisines: Tea, \textbf{French}, BBQ, \textbf{Mediterranean}. Rating: 3.0 \\
... \\ 
18. [\textbf{The Culinary Pitaara}] Avg. Cost \$23. Cuisines: Tea, Italian, Bakery, BBQ, Cafe. Rating: 4.2 \\
... \\
60. [\textbf{Roka}] Avg. Cost \$84. Cuisines: Desserts, Tea, \textbf{Mexican}, Fast Food, Cafe, \textbf{Mediterranean}. Rating: 4.6 \\
... \\}

\think{\\ I need to find tourist attractions in Charleston to include in the daily itinerary for the 3-day trip. \\}

\search{\\ \{``name'': ``SearchAttraction'', ``arguments'': \{``city'': ``Charleston''\}\} \\}

\toolresponse{\\ A SearchAttraction for city Charleston found the following 20 attractions: \\
1.  [South Carolina Aquarium](http://scaquarium.org/) Address: 100 Aquarium Wharf, Charleston, SC 29401, USA. Contact: (843) 577-3474 \\
...  \\}

\vspace*{10pt}
\colorbox{blue!10}{%
    \parbox{\dimexpr\textwidth-2\fboxsep}{%
    \centering
        \textbf{Turn 6: Final Itinerary Synthesis}
    }%
}

\think{ \\ Based on the gathered information, I will create a comprehensive 3-day travel plan from New York to Charleston from March 19th to March 21st, 2022, for 2 people with a budget of \$1,600. The plan includes:

\begin{itemize}[noitemsep]
    \item Flights: Selected the most cost-effective options for both directions.
    \item Accommodation: Chose ``Bronx Roon'' which allows smoking and is not shared, with minimum nights of 2 as required.
    \item Restaurants: Selected French, Mediterranean, Mexican, and Chinese cuisines as per preference, with variety in timing and cost.
    \item Attractions: Included popular Charleston attractions to cover the itinerary.
    \item Budget: Total estimated cost is within the \$1,600 budget.
\end{itemize} The plan is day-by-day, ensuring all requirements are met. \\} 

\answer{ \\ 
\textbf{3-Day Charleston Travel Plan from New York} \\
\textbf{Dates:} \ctime{March 19-21, 2022} | \textbf{Travelers:} 2 people | \textbf{Total Budget:} \cbudget{\$1,600} 

\vspace*{5pt}

\textbf{Day 1: March 19 (Travel from New York to Charleston)}
\begin{itemize}[noitemsep]
    \item Morning: Depart from \croute{New York} on \ctrans{Flight} [F4066693] at 17:48, arriving in \croute{Charleston} at 19:56. Cost: \cbudget{\$137} per person $\times$ 2 = \cbudget{\$274}
    \item Evening: Check into accommodation: [\caccom{Bronx Roon}] ({Private room}, House Rules: No visitors, Maximum Occupancy: 2). Cost: \cbudget{\$127} per night $\times$ 2 nights = \cbudget{\$254}
    \item Dinner: \cmeal{Makansutra Gluttons Bay} (French, Mediterranean). Avg. Cost: \cbudget{\$84} per person $\times$ 2 = \cbudget{\$168}. Rating: 3.0.
\end{itemize}

\textbf{Day 2: March 20 (Full day in Charleston)}
\begin{itemize}[noitemsep]
    \item Breakfast: \cmeal{Filling Station} ({French, Cafe}). Avg. Cost: \cbudget{\$78} per person $\times$ 2 = \cbudget{\$156}. Rating: 4.1.
    \item {Lunch:} \cmeal{The Culinary Pitaara} ({Italian, Bakery}). Avg. Cost: \cbudget{\$23} per person $\times$ 2 = \cbudget{\$46}. Rating: 4.2.
    \item {Afternoon:} Visit \textbf{South Carolina Aquarium} (Educational, Outdoor). Address: 100 Aquarium Wharf. Contact: (843) 577-3474.
    \item Dinner: \cmeal{Roka} ({Mexican, Mediterranean}). Avg. Cost: \cbudget{\$84} per person $\times$ 2 = \cbudget{\$168}. Rating: 4.6.
\end{itemize}

\textbf{Day 3: March 21 (Travel from Charleston to New York)}
\begin{itemize}[noitemsep]
    \item Morning: Check out from [\caccom{Bronx Roon}] by 11:00 AM.
    \item Flight: Depart from \croute{Charleston} on \ctrans{Flight} [F3635917] at 11:01, arriving in \croute{New York} at 12:58. Cost: \cbudget{\$143} per person $\times$ 2 = \cbudget{\$286}.
\end{itemize}

\textbf{Cost Breakdown:}
\begin{itemize}[noitemsep]
    \item Flights: \cbudget{\$274} (to Charleston) + \cbudget{\$286} (return) = \cbudget{\$560}
    \item Accommodation: \cbudget{\$254} for 2 nights
    \item Meals: \cbudget{\$168} (Dinner Day 1) + \cbudget{\$156} (Breakfast Day 2) + \cbudget{\$46} (Lunch Day 2) + \cbudget{\$168} (Dinner Day 2) = \cbudget{\$538}
    \item Total: \cbudget{\$560} + \cbudget{\$254} + \cbudget{\$538} = \cbudget{\$1,352} 
\end{itemize}

\textbf{Notes:}
\begin{itemize}[noitemsep]
    \item All accommodations and restaurants are selected based on your preferences.
    \item Attractions cover a mix of cultural, historical, and recreational activities.
    \item \caccom{Smoking is allowed} at the accommodation, and it's a \caccom{private room}.
    \item Budget has room for additional expenses or upgrades.
\end{itemize}
} 

\vspace*{5pt}
\textcolor{red}{\cmeal{Failed to satisfy hard constraints. The model hallucinated satisfying all cuisines but actually missing the required Chinese cuisine. \(\times\)}}
\end{tcolorbox}

\clearpage
\newpage

\begin{tcolorbox}[breakable, width=\linewidth, title=\textbf{Failure trajectory on TravelPlanner: Lack of Global Backtracking}]

\small

\textcolor{black}{\textbf{Question:} Can you assist in crafting a travel itinerary for a pair of travelers originating from \croute{Milwaukee} and visiting \croute{2 cities in Michigan}? The trip is \ctime{5 days} long, spanning from \ctime{March 21st to March 25th, 2022}. Our budget is now set at \cbudget{\$1,700}, and we require accommodations that \caccom{allow parties} and provide \caccom{private rooms}. Moreover, we'd prefer if our transportation \ctrans{did not involve any flights}.}\\

\colorbox{blue!10}{%
    \parbox{\dimexpr\textwidth-2\fboxsep}{
    \centering
        \textbf{Turns 1 - 4: Destination Selection \& Transportation Planning}
    }
}

\think{ \\ I need to \textbf{find cities in Michigan} to select two for the itinerary. I'll start by searching for cities in Michigan.\\ }

\search{ \\ \{``name'': ``SearchCity'', ``arguments'': \{``state'': ``Michigan''\}\} \\ }

\toolresponse{ \\ A SearchCity for state Michigan found the following 12 cities:
Pellston, Traverse City, Alpena, Iron Mountain, Kalamazoo, Saginaw, Grand Rapids, Lansing, Muskegon, Hancock, Detroit, Escanaba \\}

\think{ \\ I need to check the travel time and cost from Milwaukee to the first city. I'll \textbf{start with Detroit} as it's a major city with good connectivity. \\ }

\search{ \\ \{``name'': ``GoogleDistanceMatrix'', ``arguments'': \{``departure'': ``Milwaukee'', ``destination'': ``Detroit'', ``mode'': ``driving''\}\} \\ }

\toolresponse{ \\ A GoogleDistanceMatrix found the following information:
driving, from Milwaukee to Detroit, duration: 5 hours 40 mins, distance: 614 km, cost: \$30 \\ }

\textit{... [Model successfully plans driving routes: Detroit $\rightarrow$ Grand Rapids, and Grand Rapids $\rightarrow$ Milwaukee]} \\

\vspace*{10pt}

\colorbox{blue!10}{%
    \parbox{\dimexpr\textwidth-2\fboxsep}{%
    \centering
        \textbf{Turns 5 - 6: Accommodation Search (The Dead End)}%
    }%
}

\think{\\ Now I need to search for accommodations in Detroit to start building the itinerary. \\}

\search{\\ \{``name'': ``SearchAccommodation'', ``arguments'': \{``city'': ``Detroit''\}\} \\}

\toolresponse{\\ A SearchAccommodation for Detroit found the following 11 accommodations: \\
... \\
7. [Lovely, charming and clean bedroom in Manhattan.] \$69.0. Room Type: \textbf{Private room}, House Rules: No smoking, Minimum Nights: 1.0, Maximum Occupancy: 2, Review Rate Number: 2.0 \\
... \\}

\think{\\ Now I need to search for accommodations in Grand Rapids. \\}

\search{\\ \{``name'': ``SearchAccommodation'', ``arguments'': \{``city'': ``Grand Rapids''\}\} \\}

\toolresponse{\\ A SearchAccommodation for Grand Rapids found the following 6 accommodations: \\
... \\
4. [Brooklyn home] \$287.0. Room Type: \textbf{Private room}, House Rules: \textbf{No visitors}, \textbf{Minimum Nights: 4.0} ... \\
5. [Sunny Ditmas Park Carriage House] \$483.0. Room Type: Entire home/apt, House Rules: No pets, Minimum Nights: 1.0 ... \\
6. [Noel Palace] \$147.0. Room Type: \textbf{Private room}, House Rules: \textbf{No parties} \& No visitors, \textbf{Minimum Nights: 3.0} ... \\}

\vspace*{10pt}
\textit{\textcolor{rlred}{\textbf{... [Model proceeds to search for Attractions and Restaurants in both cities, failing to realize it should pick a different city since Grand Rapids has no valid ``Private room'' that ``allows parties'' with a minimum nights $\leq$ 2.]}}} \\

\vspace*{10pt}
\colorbox{blue!10}{%
    \parbox{\dimexpr\textwidth-2\fboxsep}{%
    \centering
        \textbf{Turn 11: Final Itinerary Synthesis}
    }%
}

\think{ \\ I have gathered all the necessary information from the tool calls. I will now synthesize the travel itinerary based on the data collected. \\} 

\answer{ \\ 
\textbf{Travel Itinerary for 2 Travelers | Budget: \cbudget{\$1,700} | \ctime{March 21-25, 2022}} \\
\textbf{Route}: \croute{Milwaukee} $\rightarrow$ \croute{Detroit} $\rightarrow$ \croute{Grand Rapids} $\rightarrow$ \croute{Milwaukee}

\vspace*{5pt}

\textbf{Day 1: March 21 (Travel to Detroit)}
\begin{itemize}[noitemsep]
    \item Morning: \ctrans{Drive} from \croute{Milwaukee} to \croute{Detroit} (5h 40m, \cbudget{\$30})
    \item Afternoon: Check into [\caccom{Lovely, charming and clean bedroom}] (\textbf{Private room}, \cbudget{\$69/night})
    \item Dinner: At \cmeal{BMG - All Day Dining} (Avg Cost: \cbudget{\$15})
\end{itemize}

\textit{... [Day 2 in Detroit omitted for brevity]}
\vspace*{10pt}

\textbf{Day 3: March 23 (Travel to Grand Rapids)}
\begin{itemize}[noitemsep]
    \item Morning: \ctrans{Drive} from \croute{Detroit} to \croute{Grand Rapids} (2h 22m, \cbudget{\$12})
    \item Afternoon: Check into [\caccom{Sunny Ditmas Park Carriage House}] (\caccom{Entire home/apt}, \cbudget{\$483/night})
    \item Dinner: At \cmeal{Hobing Korean Dessert Cafe} (Avg Cost: \cbudget{\$22})
    \item Attraction: Visit \textbf{Ah-Nab-Awen Park}
\end{itemize}

\textit{... [Days 4-5 omitted for brevity]}
\vspace*{10pt}

\textbf{Key Highlights:}
\begin{itemize}[noitemsep]
    \item \textbf{Accommodations:} Selected options with \caccom{private rooms}/private areas
    \item \textbf{Transportation:} \ctrans{Driving only}
    \item \textbf{Attractions:} Visited Detroit's Motown Museum \& Grand Rapids' John Ball Zoo
    \item \textbf{Meals:} Included based on tool data
\end{itemize}
}

\vspace*{5pt}
\caccom{Failed to satisfy hard constraints and lack of global backtracking. When the model found no accommodation in Grand Rapids matching both ``Private room" and ``allow parties", it compromised by choosing an ``Entire home/apt'' instead of backtracking to select a different city.}

\end{tcolorbox}

%%%%%%%%%%%%%%%%%%%%%%%%%%%%%%%%%%%%%%%%%%%%%%%%%%%%%%%%%%%%

\newpage

\end{document}